\documentclass[11pt]{article}
\usepackage[preprint]{acl}
\usepackage{times}
\usepackage{latexsym}
\usepackage[T1]{fontenc}
\usepackage[utf8]{inputenc}
\usepackage{microtype}
\usepackage{threeparttable}
\usepackage{amsmath,amssymb}
\usepackage{booktabs}
\usepackage{multirow}
\usepackage{graphicx}
\usepackage[dvipsnames,table]{xcolor}
\usepackage{colortbl}
\usepackage[most]{tcolorbox}
\usepackage{tikz}
\usetikzlibrary{arrows.meta,fit,calc}
\definecolor{CLRheader}{RGB}{214,232,248}    % table header fill (saturated blue)
\definecolor{CLRrow}{RGB}{241,247,253}       % alternating table row tint
\definecolor{CLRgood}{RGB}{232,245,233}      % green tint: color EICS (high)
\definecolor{CLRbad}{RGB}{255,235,238}       % red tint: spatial EICS (below BoW)
\definecolor{CLRbow}{RGB}{255,248,220}       % cream: BoW baseline rows
\definecolor{CLRref}{RGB}{245,245,245}       % grey: external reference baselines
\definecolor{CLRfind}{RGB}{31,119,180}       % finding-box frame / title bg
\definecolor{CLRfindbg}{RGB}{235,244,252}    % finding-box body background
\definecolor{CLRheatok}{RGB}{46,140,70}      % confusion tables: diagonal (correct)
\definecolor{CLRheaterr}{RGB}{214,60,50}     % confusion tables: off-diagonal (error)
\definecolor{CLRwarn}{RGB}{180,30,30}        % alarming result: text color only

\tcbset{
  findingbox/.style={
    enhanced,
    colback=CLRfindbg,
    colframe=CLRfind,
    fonttitle=\bfseries\small,
    coltitle=white,
    attach boxed title to top left={yshift=-2mm, xshift=4mm},
    boxed title style={colback=CLRfind, rounded corners},
    rounded corners,
    boxrule=0.8pt,
    left=4pt, right=4pt, top=6pt, bottom=4pt,
    before upper={\small},
  }
}
\newcommand{\finding}[2]{%
  \begin{tcolorbox}[findingbox, title={Finding #1}]%
    #2%
  \end{tcolorbox}%
}

\title{When VLMs Destroy What They Know:\\
Spatial Representations Fall Below Lexical Priors}

\author{
  Shubh Garg \\
  Thapar Institute of Engineering and Technology \\
  \texttt{shubhgarg265@gmail.com} \And
  Chen Shani \\
  Tel Aviv University \\
  \texttt{cshani@tauex.tau.ac.il}
}

\begin{document}
\maketitle

% ── ABSTRACT ─────────────────────────────────────────────────────────────────
\begin{abstract}
Vision-Language Models (VLMs) increasingly analyze and reason about images, yet it remains unclear whether their internal representations faithfully encode the structured relations required for such reasoning. We introduce EICS (\textbf{E}valuation-\textbf{I}nvariant \textbf{C}oncept \textbf{S}tability), a metric for testing whether representations consistently encode concepts under controlled one-concept interventions, and CASP-VL (\textbf{CA}usally controlled \textbf{S}patial \textbf{P}erturbations for \textbf{V}ision-\textbf{L}anguage models), a benchmark of causally controlled spatial perturbations derived from Visual Genome scene graphs. Across multiple VLM architectures, we find that spatial relations are represented substantially less faithfully than lexical information, even when the same models score well above chance on compositional benchmarks. Further analysis shows that the relation is identifiable at the predicate's own tokens but is lost when the sentence is aggregated into a global representation, whether through the \texttt{[EOS]} token or by mean pooling. These findings reveal a systematic failure invisible to output-based evaluation: VLMs do not preserve, in the representation used downstream, relational information that is present in their own input. Code and data:
\url{https://github.com/Shubhf/CASP-VL}.
\end{abstract}

% ── 1. INTRODUCTION ──────────────────────────────────────────────────────────
\section{Introduction}
\label{sec:intro}

When a Vision-Language Model (VLM) correctly answers ``is the cup on the table?'', it is unclear whether it represents the spatial relation between the cup and table, or instead relies on statistical associations between the words in the input. Prior work has exposed such failures by testing VLMs on compositional and order-sensitive tasks~\citep{yuksekgonul2023aro,hsieh2023sugarcrepe,zhao2024mmcomposition}. These studies reveal important limitations in VLM behavior, but they leave open a more fundamental question: \emph{do VLMs internally represent the concepts required for these tasks?}

This distinction matters because behavioral success does not necessarily imply faithful internal representations. A model may produce the correct answer while relying on superficial cues that fail when those cues change. Conversely, a model may encode relevant information internally even when its final prediction fails. For systems expected to reason about spatial structure under changing conditions, understanding what is represented is therefore complementary to measuring what the model predicts.

We address this question through \emph{concept stability}: whether an internal representation changes in a predictable, decodable direction when exactly one underlying concept is changed and everything else is held fixed. We introduce CASP-VL, a benchmark of causally controlled spatial perturbations derived from Visual Genome~\citep{krishna2017visualgenome}, and EICS, a metric for measuring representational stability under these interventions. We compare VLM representations against a simple BoW lexical baseline, which captures information available from token identities alone without contextual encoding.

Our results reveal that VLMs represent spatial relations \textit{less faithfully} than a simple BoW, even when they score well above chance on compositional benchmarks. In CLIP, the relation is identifiable at the predicate's own tokens but is largely lost in the pooled sentence representation. Thus, the failure is not simply that VLMs cannot see the relation word: it is identifiable locally, but is not preserved in the representation used for downstream decisions. More broadly, our findings show that output-based evaluation can miss systematic failures in how VLMs represent structured concepts.

We make three contributions:
\textbf{(1)~CASP-VL}: a benchmark of controlled one-concept edits for testing whether VLM representations encode spatial relations. \textbf{(2)~EICS}: a metric for measuring representational faithfulness under concept interventions, separating conceptual encoding from surface-level lexical cues. \textbf{(3)~Empirical analysis}: a systematic evaluation across layers and VLM architectures showing that spatial structure is represented less faithfully than lexical information, and identifying where spatial information is lost.

% ── 2. RELATED WORK ──────────────────────────────────────────────────────────
\section{Related Work}
\label{sec:related}

\begin{table}[t]
\centering
\caption{\textbf{Prior compositional benchmarks score model outputs; EICS probes the representation, against a lexical baseline, at every layer.}
ARO~\citep{yuksekgonul2023aro}, SugarCrepe~\citep{hsieh2023sugarcrepe},
Winoground~\citep{thrush2022winoground}, What'sUp~\citep{kamath2023whatsup}; \emph{Ours} = CASP-VL
+ EICS. \emph{Level}: what is evaluated (model output or internal representation).
\emph{Lexical}: a baseline that sees only the words (text-only scorers for SugarCrepe, the
BoW probe for ours). \emph{Layers}: results reported per layer.}
\label{tab:positioning}
\small
\setlength{\tabcolsep}{2.6pt}
\begin{tabular}{llccc}
\toprule
\rowcolor{CLRheader}
Benchmark & Edit & Level & Lexical & Layers \\
\midrule
ARO        & word swaps     & output & $\times$   & $\times$ \\
SugarCrepe & LLM negatives  & output & \checkmark & $\times$ \\
Winoground & word order     & output & $\times$   & $\times$ \\
What'sUp   & object layout  & output & $\times$   & $\times$ \\
\midrule
\rowcolor{CLRgood}\textbf{Ours} & one concept & \textbf{repr.} & \checkmark & \checkmark \\
\bottomrule
\end{tabular}
\end{table}

\paragraph{Compositional benchmarks.}
\citet{yuksekgonul2023aro} showed CLIP behaves like a BoW model on
relational expressions. \citet{thrush2022winoground} demonstrated failure on
visio-linguistic compositionality even with familiar words and images.
\citet{hsieh2023sugarcrepe} showed prior benchmarks' hard negatives are
exploitable through language priors, motivating our causal perturbation design.
All of these score model outputs (Table~\ref{tab:positioning}).

\paragraph{Training-time compositional methods.}
Several methods improve compositional behavior by injecting structure during training. SGVL~\citep{herzig2023incorporating} learns from scene graphs, adding a scene-graph component to the image encoder and generating fine-grained captions from the graphs; CS-CLIP~\citep{jeong2026cs} uses scene graphs to construct structured hard negatives; MACCO~\citep{li2026cross} masks compositional concepts in one modality and reconstructs them from the other. These methods are evaluated by whether the correct caption outscores a hard negative. EICS asks a complementary question of the text encoder alone: whether its representation changes in a consistent, decodable direction when only the relation changes. Whether these methods also raise EICS is an open question that our released code makes straightforward to test.

\paragraph{Spatial reasoning.}
\citet{kamath2023whatsup} showed VLMs struggle with basic spatial terms.
\citet{chen2024spatialvlm} proposed explicit spatial supervision.
\citet{chen2025spatial_attention} traced spatial reasoning errors to
attention directed at irrelevant image regions. Our work adds a
representation-level account: spatial relations are suppressed \emph{below}
what input tokens alone predict.

\paragraph{Probing and causal methods.}
\citet{belinkov2022probing} established probing classifiers for characterizing
encoded linguistic properties. \citet{wani2025same} showed that VLMs can keep
their answers unchanged while their internal representations drift
substantially, motivating representation-level evaluation.
\citet{kang2026linear} found that VLMs bind object locations to textual
activations through linear spatial IDs that causally mediate model beliefs.
\citet{zhang2025cfvlm} fine-tune VLMs on counterfactual image--text pairs and
\citet{parascandolo2025cogt} impose causal dependency structure on caption
decoding; both target output-level behavior, whereas EICS asks the prior
question at the representation level.

\paragraph{Information Bottleneck.}
\citet{tishby2000information} introduced the IB framework for optimal
compression. \citet{yang2025exploring} use the IB lens to trace how LLMs
compress and then predict across layers, and \citet{shani2026tokens} show
that LLMs favor aggressive statistical compression over fine-grained semantic
distinctions. Our layer-wise and predicate-token results give a
representation-level view of such compression in VLMs: spatial signal that is
present at the predicate token is largely lost in the pooled representation.

% ── 3. CASP-VL AND EICS ──────────────────────────────────────────────────────
\section{CASP-VL and EICS}
\label{sec:benchmark}

\begin{figure*}[t]
  \centering
  \resizebox{\textwidth}{!}{\input{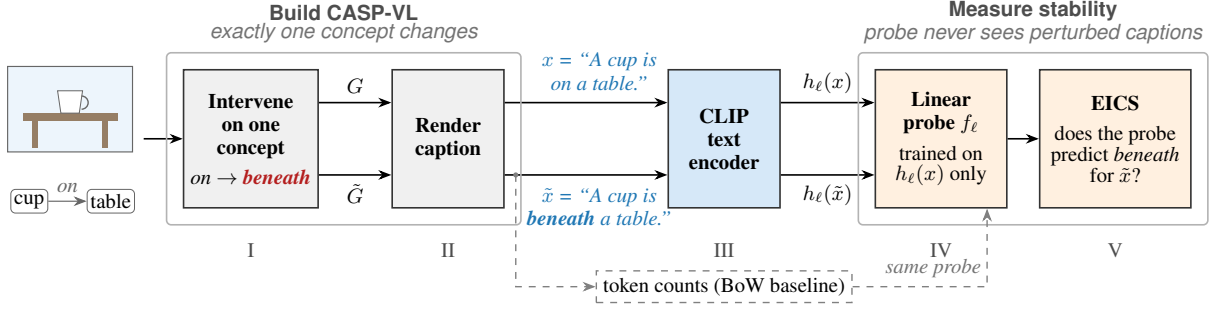}}
  \caption{\textbf{EICS asks whether a representation follows a one-concept change in its input.}
  (I)~We edit one concept in a Visual Genome scene graph $G$, giving $\tilde G$. (II)~Both graphs are
  rendered as captions $x$ and $\tilde x$ from one template. (III)~The text encoder maps each caption
  to a hidden state. (IV)~A linear probe trained only on original captions (V)~is tested on the
  perturbed ones: EICS is how often it predicts the new label, minus chance. The same probe on token
  counts (dashed) gives the BoW lexical baseline.}
  \label{fig:pipeline}
\end{figure*}

\subsection{Benchmark Construction}
\label{sec:construction}

We introduce \textbf{CASP-VL} (\textbf{CA}usally controlled \textbf{S}patial \textbf{P}erturbations for \textbf{V}ision-\textbf{L}anguage models), a benchmark designed to test whether VLM representations track changes in visual concepts. We construct controlled image-caption triplets in which exactly one concept is changed, allowing representational responses to be evaluated independently of standard task accuracy.

\paragraph{Source and perturbations.}
We build on Visual Genome~\citep{krishna2017visualgenome}, which contains 108,077 images with dense scene-graph annotations. We intervene on the scene graphs rather than the captions, ensuring that each triplet differs in exactly one concept. We consider two intervention types: \textbf{attribute flips}, which replace an attribute with another of the same type (e.g., \emph{red} $\rightarrow$ \emph{blue}), and \textbf{relation flips}, which replace a spatial predicate with a different one drawn from a fixed list of alternatives (e.g., \emph{on} $\rightarrow$ \emph{beneath}). A caption renders the first five relationships of the image's scene graph with the template \emph{``A [attr] [subj] is [pred] a [attr] [obj].''}, one sentence each, using each object's first attribute. A relation flip changes one sentence; a color flip changes the color of one object instance, and therefore every sentence that mentions that instance. Because Visual Genome can annotate the same relationship for several instances, the old label can survive elsewhere in the perturbed caption; we analyze this in Appendix~\ref{app:errors}. Sentences can repeat when Visual Genome annotates the same relationship more than once, and one image can contribute several triplets (an example is in Appendix~\ref{app:reeval}).

\paragraph{Notation.}
Each item $i$ starts from an image $I_i$ and its scene graph $G_i$. Editing exactly one concept gives $\tilde G_i$ (stage~I in Figure~\ref{fig:pipeline}), and the template renders the two graphs as an original caption $x_i$ and a perturbed caption $\tilde x_i$ (stage~II). The edited concept has label $y_i$ in $x_i$ and $\tilde y_i$ in $\tilde x_i$; for ``A cup is on a table'' $\rightarrow$ ``A cup is beneath a table'', $y_i=$ \emph{on} and $\tilde y_i=$ \emph{beneath}. We call $(I_i, x_i, \tilde x_i)$ a \emph{triplet}, and drop $i$ when it is clear, as in the figure. The edits are interventions in the sense that one variable changes while everything else is held fixed, but they act on the caption only: $I_i$ is never modified, so $\tilde x_i$ is a counterfactual description of the same image.

\paragraph{Scale and balance.}
CASP-VL contains 2,400 color triplets (200 per class across 12 classes) and 1,491 spatial triplets (up to 100 per class across 16 classes); Appendix~\ref{app:stats} gives full statistics and every edit. We additionally construct 4,200 cross-category triplets for the taxonomy analysis in Appendix~\ref{app:taxonomy}. All subsets use balanced class distributions.

\subsection{EICS Metric}
\label{sec:eics}

\begin{figure*}[t]
  \centering
  \includegraphics[width=0.8\textwidth]{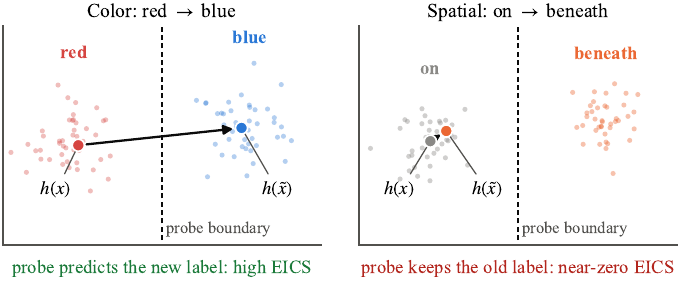}
  \caption{\textbf{A color flip moves the representation across the probe's decision boundary; a spatial flip does not.}
  Schematic. Left: flipping ``red'' to ``blue'' moves the representation into the blue cluster, so a
  probe trained on original captions predicts the new label (high EICS). Right: flipping ``on'' to
  ``beneath'' leaves the representation inside the ``on'' cluster, so the probe still predicts the
  old label (near-zero EICS).}
  \label{fig:eics_concept}
\end{figure*}

We introduce \textbf{EICS} (Evaluation-Invariant Concept Stability) to measure whether a representation tracks a concept intervention beyond what can be inferred from surface lexical statistics. Given a probe trained only on the original captions, EICS evaluates whether it predicts the changed label for the corresponding perturbed captions:
\begin{equation}
\text{EICS}(\ell, c) = \Pr\!\left[f_\ell\!\left(h_\ell(\tilde{x}_i)\right) = \tilde{y}_i\right] - \frac{1}{|\mathcal{C}|}
\label{eq:eics}
\end{equation}
\noindent Here, $h_\ell(x)$ denotes the \texttt{[EOS]}-pooled hidden state of the text encoder at layer $\ell$ (stage~III in Figure~\ref{fig:pipeline}), $f_\ell$ is a logistic-regression probe trained on $h_\ell(x_i)$ with labels $y_i$ only (stage~IV), and $|\mathcal{C}|$ is the number of concept classes. EICS (stage~V) asks how often the probe assigns the \emph{new} label $\tilde y_i$ to the perturbed caption. EICS is zero at chance and increases as the representation more reliably tracks the intervened concept (Figure~\ref{fig:eics_concept}). We report 95\% bootstrap confidence intervals over 1,000 resamples.

As a reference point, we apply the same procedure to normalized word-count vectors, yielding a \textbf{BoW lexical baseline}. This baseline captures information available from word identity alone, without contextual encoding, and provides a measure of how well a representation preserves lexical information already present in its input (details in Appendix~\ref{app:bow}).

% ── 4. EXPERIMENTS ───────────────────────────────────────────────────────────
\section{Experiments}
\label{sec:experiments}

We ask whether VLM representations faithfully encode spatial relations, and if not, where and how this information is lost. We proceed in four steps. First, we test whether representations respond to controlled changes in spatial relations at all (\S\ref{sec:rq1}). Second, we use EICS to test whether these changes are encoded in a way that supports stable decoding, comparing spatial structure with lexical information across layers and architectures (\S\ref{sec:rq2}). Third, we localize the failure by testing argument order and representations at the predicate token rather than the pooled representation (\S\ref{sec:rq3}). Finally, we contrast EICS with accuracy on established compositional benchmarks and their baselines (\S\ref{sec:rq4}). Additional analyses of visual grounding (Appendix~\ref{app:grounding}), probe robustness (Appendices~\ref{app:bow} and~\ref{app:probe}), relation-type geometry (Appendix~\ref{app:taxonomy}), caption style (Appendix~\ref{app:paraphrase}), and model scale (Appendix~\ref{app:larger}) are reported in the appendix, together with an error analysis (Appendix~\ref{app:errors}). Appendix~\ref{app:faq} answers the most likely objections in brief.

\begin{figure}[t]
  \centering
  \includegraphics[width=\columnwidth]{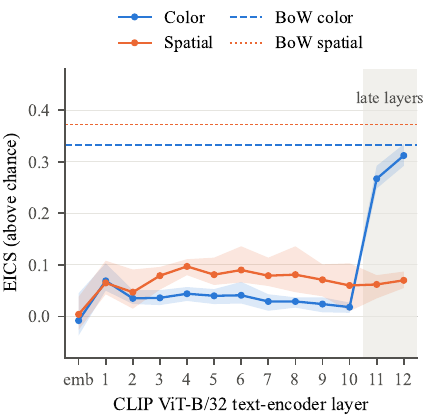}
  \caption{\textbf{Color becomes decodable in the last two layers; spatial relations stay far below the lexical baseline at every layer.}
  EICS by layer for CLIP ViT-B/32 (12 color classes, 16 relation classes), averaged over five
  image-grouped splits; bands span the lowest and highest split. Dashed and dotted lines: the BoW
  baseline for each concept, averaged over the same splits.}
  \label{fig:layerwise}
\end{figure}

\paragraph{Setup.}
We use CLIP ViT-B/32~\citep{radford2021clip} as our primary model and evaluate SigLIP-B/16~\citep{zhai2023siglip} and LLaVA-1.5-7B~\citep{liu2024llava} to test whether findings generalize across architectures and training objectives. Unless otherwise stated, we extract hidden states at \texttt{[EOS]} across all 13 CLIP layers. We train $\ell_2$-regularized logistic regression probes on the original captions of 80\% of the triplets and compute EICS (Eq.~\ref{eq:eics}) on the perturbed captions of the held-out 20\%. The split is grouped by image, so no image contributes to both sides, and the probe never sees a perturbed caption. Some relation flips are one-directional antonyms (\emph{near} $\rightarrow$ \emph{far from}) whose new label never occurs on an original caption; since the probe cannot predict a class it was never trained on, we exclude these unscorable test flips. We repeat this split with five seeds (42, 1, 2, 3, 4) and report the mean and standard deviation across splits for the CLIP ViT-B/32 results in this section (Figure~\ref{fig:layerwise}, Table~\ref{tab:predtoken}); analyses run on one split use seed 42 (482 color and 242 spatial test triplets) and report 95\% bootstrap intervals over its held-out triplets (Appendix~\ref{app:reeval}). Only the first BoW robustness check (Appendix~\ref{app:bow}) was run earlier, with a triplet-level split that keeps the unscorable flips; its absolute values, BoW's in particular, are lower, but every comparison between CLIP and BoW points the same way. Different analyses use different data: the main results use the 3{,}891 CASP-VL triplets; the argument-order test (\S\ref{sec:rq3}) uses a separate set of single-sentence caption pairs; the taxonomy analysis (Appendix~\ref{app:taxonomy}) a separate set of 4{,}200 cross-category triplets; and visual grounding (Appendix~\ref{app:grounding}) a 500-triplet sample. All are drawn from Visual Genome, so the separate sets can share images with CASP-VL, but no analysis mixes triplets across sets. Hardware, software, and all settings needed to reproduce the results are listed in Appendix~\ref{app:repro}.

\subsection{Do VLM Representations Respond to Spatial Changes?}
\label{sec:rq1}

We first ask a basic question: \emph{does a controlled change in a spatial relation produce a corresponding change in the representation?} This tests representational sensitivity before asking whether the changed concept can be decoded.

We compare the cosine similarity between original and perturbed CLIP text representations. Despite the edit, representations remain highly similar, and most of all for spatial relations: at L12, the mean cosine similarity between original and perturbed \texttt{[EOS]} states is $0.992$ for relation flips, against $0.943$ for color flips (Table~\ref{tab:cosine}, Appendix~\ref{app:cosine}). Thus, CLIP's text representation is substantially less sensitive to changes in spatial relations than to comparable attribute changes.

\finding{1}{Spatial flips barely move CLIP's text representation; color flips do.}

This establishes a basic representational failure, but similarity alone cannot tell us whether the small change that does occur contains decodable information about the intervened concept. We therefore turn to EICS.

\subsection{Do Representations Faithfully Encode Spatial Relations?}
\label{sec:rq2}

We next ask the central question of the paper: \emph{when a spatial relation changes, does the representation change in a direction that reliably tracks the new relation?} EICS addresses this by training probes only on the original captions and evaluating them on the corresponding perturbed captions.

\paragraph{Layer-wise analysis.}
Figure~\ref{fig:layerwise} shows EICS across the CLIP hierarchy. Averaged over five splits, color is barely decodable through L10 and then rises sharply to $+0.312 \pm 0.019$ at L12, 94\% of its BoW baseline ($+0.332 \pm 0.016$). Spatial EICS stays small at every layer: it peaks mid-network ($+0.097$ at L04) and ends at $+0.070 \pm 0.012$ at L12, under a fifth (19\%) of its BoW baseline ($+0.372 \pm 0.022$). The ranking holds on every split: CLIP keeps 81--102\% of the BoW signal for color and 16--22\% for spatial relations (Appendix~\ref{app:reeval}). Some perturbed color captions still mention the old color elsewhere; without them (split 42), color EICS ($+0.391$) exceeds BoW ($+0.326$), while spatial EICS stays at a fifth of BoW ($+0.074$ vs.\ $+0.366$; Appendix~\ref{app:errors}). The same pattern holds for wider CLIP text encoders (ViT-L/14 and ViT-L/14@336; Appendix~\ref{app:larger}) and when the template captions are replaced by human-written Visual Genome descriptions or fluent LLM paraphrases (Appendix~\ref{app:paraphrase}). Reducing each caption to the edited sentence narrows the spatial gap without closing it: CLIP then keeps 98\% of the BoW signal for color but 52\% for spatial relations (split 42; Appendix~\ref{app:paraphrase}).

The comparison with BoW is important: the model has access to the relevant lexical information in its input, yet its contextual representation preserves substantially less of that information for spatial relations. Thus, the result is not simply that spatial concepts are difficult to decode; the representation preserves less spatial information than a representation that contains no contextual encoding at all.

\paragraph{Across architectures.}
We next ask whether this pattern is specific to CLIP's contrastive objective. Table~\ref{tab:sota} reports color and spatial EICS for CLIP, SigLIP, and LLaVA under the same protocol, with confidence intervals in Table~\ref{tab:models_ci} (Appendix~\ref{app:benchmarks}). No model's spatial EICS comes near the BoW spatial baseline ($+0.359$): CLIP ViT-B/32 reaches $+0.066$, CLIP ViT-L/14 $+0.053$, SigLIP $+0.038$, and LLaVA $+0.115$. For the three contrastive models, the 95\% CIs for color and spatial EICS do not overlap, so the color--spatial gap is statistically clear. LLaVA's last prompt token keeps little of either concept: color ($+0.143$ [0.107, 0.182]) and spatial EICS ($+0.115$ [0.070, 0.165]) overlap and both stay far below BoW, although LLaVA's own predicate tokens still identify the edited word (EICS of at least $+0.86$ for both concepts). We therefore do not claim a color--spatial gap for LLaVA; its spatial EICS stays below the BoW baseline of the same split, whose CI it does not overlap ([0.305, 0.417]). SigLIP illustrates that greater representational sensitivity does not necessarily produce more faithful spatial encoding.

Together, these results establish that the spatial encoding failure is not specific to a single layer or architecture. VLM representations consistently preserve spatial structure less faithfully than lexical information.

\finding{2}{For spatial relations, no VLM we test keeps more information than a bag of words.}

\subsection{Is Spatial Structure Encoded at All?}
\label{sec:rq3}

The preceding analysis raises an important alternative explanation: perhaps the model does not represent spatial relations anywhere. We test this directly using two complementary analyses.

\begin{table}[t]
\centering
\caption{\textbf{No sentence-level pooling keeps the spatial relation, while the predicate's own tokens identify it almost perfectly.}
CLIP ViT-B/32 at L12, mean $\pm$ s.d.\ over five image-grouped splits. \emph{Predicate tokens}: mean over the edited word's tokens; its ceiling is
$1-1/|\mathcal{C}|$ (0.917 for color, 0.938 for spatial). \emph{Orig.\ acc}: probe accuracy on
original captions. Number of classes in parentheses; every value has its 95\% CI above zero on every split (split-42 CIs in
Table~\ref{tab:reeval}).}
\label{tab:predtoken}
\small
\setlength{\tabcolsep}{5pt}
\begin{tabular}{lcc}
\toprule
\rowcolor{CLRheader}
CLIP B/32, L12 & Color (12) & Spatial (16) \\
\midrule
\texttt{[EOS]}         & \cellcolor{CLRgood}$+$0.312 $\pm$ .019 & \cellcolor{CLRbad}$+$0.070 $\pm$ .012 \\
Mean pooling           & \cellcolor{CLRgood}$+$0.309 $\pm$ .025 & \cellcolor{CLRbad}$+$0.093 $\pm$ .020 \\
Predicate tokens       & $+$0.917 $\pm$ .000 & $+$0.930 $\pm$ .004 \\
\rowcolor{CLRbow}BoW   & $+$0.332 $\pm$ .016 & $+$0.372 $\pm$ .022 \\
\midrule
Orig.\ acc, CLIP/BoW & 0.65 / 0.66 & 0.24 / 0.63 \\
\bottomrule
\end{tabular}
\end{table}

\paragraph{Argument-order sensitivity.}
We first construct a particularly stringent test of structural encoding by swapping the subject and object while keeping the predicate and all tokens fixed (e.g., ``cat on mat'' $\rightarrow$ ``mat on cat''). Because the token multiset is unchanged, a BoW representation cannot distinguish the two sentences (accuracy $0.500$ by construction).

We render 1,280 single-sentence captions (80 for each of the 16 spatial predicates) and their argument-swapped versions. Both captions of a pair, and all captions from one image, go to the same side of an 80/20 split, and a binary probe on CLIP's \texttt{[EOS]} states learns to tell original from swapped captions. At L12 it reaches $0.574$ accuracy on held-out captions (95\% CI $[0.535,0.610]$), above the $0.500$ that BoW gets by construction; across layers it peaks at $0.602$ (L05). Without the symmetric predicates \emph{near}, \emph{beside} and \emph{next to}, the L12 accuracy is $0.554$ $[0.513,0.598]$. CLIP therefore keeps a weak trace of who is where, far weaker than its trace of the edited word itself. Even this trace is an upper bound on order encoding: the annotated order is usually the plausible one (\emph{``cup on table''} vs.\ \emph{``table on cup''}), so part of it may reflect plausibility rather than syntax.

\paragraph{Predicate-token probing.}
We next ask whether spatial information may exist locally but be lost when representations are pooled. We therefore probe the hidden state at the predicate token rather than \texttt{[EOS]} (Table~\ref{tab:predtoken}). For multi-word predicates such as \emph{on top of}, we average over all of the predicate's tokens.

The difference is large: pooling over the predicate's tokens reaches $+0.930$, close to its ceiling of $0.938$, against $+0.070$ at \texttt{[EOS]}. The predicate's identity is therefore available at its own position but is not carried into the pooled representation used for downstream similarity and classification.

These results refine the main finding. VLMs do not simply fail to see the relation word. Rather, the relation is identifiable locally but is substantially weakened or discarded by the aggregation used downstream.

\paragraph{Pooling.}
Mean pooling over all tokens does not recover the relation either: spatial EICS rises only from $+0.070$ to $+0.093$, far below BoW ($+0.372$), while color stays close to BoW under both poolings ($+0.312$ and $+0.309$). The loss is therefore not specific to the \texttt{[EOS]} token: neither summary of the full caption keeps the relation. For a single sentence, mean pooling keeps more ($+0.702$ vs.\ $+0.487$ at \texttt{[EOS]}, BoW $+0.938$), so the loss is largest at \texttt{[EOS]} (Appendix~\ref{app:paraphrase}). The probe struggles even before any perturbation, classifying original captions with 24\% accuracy from CLIP against 63\% from BoW (chance 6\%). Because predicate-token pooling reaches its ceiling for color as well, we read it as the edited word being identifiable at its own position, not as evidence of relational structure; the argument-order test addresses structure.

\finding{3}{The relation is identifiable at the predicate token, but pooled representations lose much of it.}

\begin{table}[t]
\centering
\caption{\textbf{Benchmark accuracy does not predict representational faithfulness: SigLIP has the best SugarCrepe relation accuracy and the worst spatial EICS.}
\emph{ARO}: VG-Relation~\citep{yuksekgonul2023aro};
\emph{SC Rel} / \emph{SC Swap}: SugarCrepe \texttt{replace\_rel} / \texttt{swap\_obj}~\citep{hsieh2023sugarcrepe}.
All three are two-way choices. EICS is measured on CASP-VL under the main protocol (split 42; Section~\ref{sec:experiments}, Setup), at the final layer of each model: L12 of the CLIP and SigLIP text encoders, and L31 of LLaVA's language model at the last prompt token (Appendix~\ref{app:repro}); all ARO and SugarCrepe subtasks are in
Appendix~\ref{app:benchmarks}.
We computed all accuracies with a single script, scoring each caption by its image--text similarity,
rather than copying them from prior papers. We report accuracy over all examples; this is why our CLIP
ViT-B/32 ARO score differs from the 59.0\% of \citet{yuksekgonul2023aro}, which is macro-averaged over relations.
For reference, \citet{hsieh2023sugarcrepe} report 97.0\% / 99.0\% for humans and at most 50.0\% for
text-only models on SC Rel / SC Swap.
BoW is shown for the CLIP split; each model's own BoW is in Table~\ref{tab:models_ci}. * = 95\% CI above zero; $\ddagger$ = generative model, contrastive
scoring not applicable; n/a = not applicable.}
\label{tab:sota}
\small
\setlength{\tabcolsep}{2.4pt}
\begin{tabular}{lccccc}
\toprule
\rowcolor{CLRheader}
& \multicolumn{3}{c}{\textbf{Accuracy (\%)}} & \multicolumn{2}{c}{\textbf{EICS}} \\
\cmidrule(lr){2-4}\cmidrule(lr){5-6}
\rowcolor{CLRheader}
Model & ARO & SC Rel & SC Swap & Color & Spatial \\
\midrule
\rowcolor{CLRref}Chance   & 50.0 & 50.0 & 50.0 & 0.000 & 0.000 \\
\rowcolor{CLRbow}BoW      & n/a  & n/a  & n/a  & $+$0.359* & $+$0.359* \\
\midrule
\rowcolor{CLRrow}CLIP B/32 & 51.8 & 69.3 & 61.5 & $+$0.292* & \textcolor{CLRwarn}{$+$0.066*} \\
CLIP L/14                  & 53.3 & 65.3 & 61.1 & $+$0.303* & \textcolor{CLRwarn}{$+$0.053*} \\
\rowcolor{CLRrow}SigLIP B/16 & 46.3 & 70.6 & 60.3 & $+$0.339* & \textcolor{CLRwarn}{$+$0.038*} \\
LLaVA-1.5                  & n/a$^\ddagger$ & n/a$^\ddagger$ & n/a$^\ddagger$ & $+$0.143* & \textcolor{CLRwarn}{$+$0.115*} \\
\bottomrule
\end{tabular}
\end{table}

\subsection{Does Benchmark Accuracy Reflect Representational Faithfulness?}
\label{sec:rq4}

A natural question is whether models that score higher on behavioral benchmarks also show higher EICS. Table~\ref{tab:sota} places both views side by side, together with chance and the BoW baseline. The two measures dissociate. CLIP ViT-B/32 scores 69.3\% on the SugarCrepe relation split, well above chance and the text-only baselines of \citet{hsieh2023sugarcrepe} (49.4\% and 50.0\%), yet its spatial EICS is $+0.070$, under a fifth of the BoW baseline of $+0.372$. Argument order is weak in both views: the three contrastive models pick the correct caption on 60--62\% of SugarCrepe's argument swaps, and a probe on CLIP's representation separates original from swapped captions with $0.574$ accuracy (\S\ref{sec:rq3}).

SigLIP is the clearest case (Table~\ref{tab:sota}): it has the highest SugarCrepe relation accuracy of the three contrastive models (70.6\%) and the lowest spatial EICS ($+$0.038, with a CI reaching down to zero). The two benchmarks do not even agree with each other: the same model is below chance on ARO (46.3\%). LLaVA reaches the highest spatial EICS ($+$0.115), still less than a third of the lexical prior. Output-based evaluation therefore misses the representational failure exposed by EICS.

% ── 5. ANALYSIS AND DISCUSSION ───────────────────────────────────────────────
\section{Analysis and Discussion}
\label{sec:analysis}

The experiments establish a consistent gap between spatial and lexical encoding. We next ask what this gap means, whether it reflects a failure of relational structure specifically, and what might drive the observed representation geometry.

\paragraph{VLMs do not exceed lexical priors.}
A useful reference point is the information already available in the input tokens. A probe on word counts achieves EICS of $+0.332$ for color and $+0.372$ for spatial relations. CLIP at L12 retains most of the color signal ($+0.312$) but little of the spatial signal ($+0.070$). Thus, contextual encoding does not improve on the lexical signal available in the input for spatial relations; instead, the final representation preserves substantially less of it. This pattern is not explained by the choice of linear probe: with a two-hidden-layer MLP, spatial EICS at L12 is $+0.086$ (split 42), and at no layer does it reach the linear BoW baseline (Appendix~\ref{app:probe}).

\paragraph{Spatial encoding loses the relation and keeps little of its structure.}
The two tests probe different aspects of relational structure. The first asks whether the representation tracks \emph{which relation} holds: spatial EICS is $+0.070$, far below the BoW baseline of $+0.372$, so it mostly does not. The second asks whether it tracks \emph{who stands in that relation to whom}: with the predicate and token multiset fixed, CLIP separates original from swapped captions with $0.574$ accuracy (\S\ref{sec:rq3}), a weak trace of order that is not enough to recover the relation. Color shows neither failure, suggesting that its encoding can be supported by lexical co-occurrence, whereas spatial relations require structural information that is not reliably preserved.

\paragraph{Movement does not imply faithfulness.}
Spatial edits barely move either text representation: the mean cosine similarity between original and perturbed states at the last layer is $0.992$ for CLIP and $0.984$ for SigLIP, against $0.943$ and $0.956$ for color edits (main protocol). SigLIP's representation moves twice as far under spatial edits ($1-\cos$ of $0.016$ vs.\ $0.008$), yet its spatial EICS is no higher ($+0.038$, CI reaching zero, vs.\ $+0.066$ for CLIP on the same split). How far a representation moves therefore says little about whether it moves toward the new relation, which is what EICS measures.

\paragraph{Visual grounding does not resolve the failure.}
One might expect access to the image to supply the spatial information missing from the text representation. We therefore compare image--caption similarity before and after each intervention (Appendix~\ref{app:grounding}). Visual grounding improves discrimination for color and size changes, but not for spatial relations: after a spatial flip, the image prefers the original caption in only 34\% of cases (Table~\ref{tab:exp2}). Thus, incorporating the visual modality does not reliably recover the intended spatial relation.

\paragraph{Other relation types collapse onto spatial ones.}
A three-way probe separating spatial, part-whole, and action relations shows the same geometry from another angle (Appendix~\ref{app:taxonomy}). On unperturbed captions it already labels most part-whole relations as spatial, and editing a spatial relation into a part-whole or action one almost never changes its prediction, while edits in the other direction do. Spatial predicates therefore occupy a broad region of the representation that absorbs other relation types, rather than a set of separable directions.

\paragraph{What would it take to fix spatial encoding?}
Together, these analyses suggest that improving spatial encoding requires more than increasing representational sensitivity or changing the contrastive objective. Two directions emerge. First, \emph{structural supervision} that explicitly penalizes argument-order errors, such as order-perturbed hard negatives~\citep{yuksekgonul2023aro} or dense spatial supervision~\citep{chen2024spatialvlm}, may force the model to preserve subject--predicate--object structure rather than collapsing to lexical co-occurrence. Second, \emph{alternative pooling mechanisms} may help, but not trivially: mean pooling over all tokens does not recover the relation (\S\ref{sec:rq3}), so a useful pooling would have to weight the predicate and its arguments, for example through learned attention. Objective changes alone, as SigLIP demonstrates, are insufficient.

% ── 6. CONCLUSION ────────────────────────────────────────────────────────────
\section{Conclusion}
\label{sec:conclusion}

We introduced EICS and CASP-VL to measure representational faithfulness in VLMs. Color information is recoverable from late-layer representations (\textbf{EICS $+$0.312}, close to the BoW baseline of $+$0.332), whereas spatial relations are not encoded in a comparably stable, decodable form at the \texttt{[EOS]} position used by downstream applications (\textbf{EICS $+$0.070, under a fifth of its BoW baseline}). The predicate's own tokens identify the relation almost perfectly (\textbf{EICS $+$0.930}), but \texttt{[EOS]} and mean pooling lose much of it. Visual grounding and objective changes do not resolve this failure, while the taxonomy analysis shows that other relation types collapse onto the spatial region. Together, these findings expose a form of representational failure that output-based evaluation can miss, with implications for how grounded VLMs are evaluated and designed.

% ── LIMITATIONS ──────────────────────────────────────────────────────────────
\section*{Limitations}
\label{sec:limitations}

\paragraph{Scope of models evaluated.}
Our primary analysis uses CLIP ViT-B/32, with SigLIP-B/16 and LLaVA-1.5-7B as cross-architecture comparisons. While the spatial encoding failure is consistent across these three models, we do not evaluate larger or more recent VLMs (e.g., CLIP ViT-L/14@336 is included only in the model-scaling analysis of Appendix~\ref{app:larger}, and models trained on richer spatial supervision such as SpatialVLM~\citep{chen2024spatialvlm} are not probed). It is possible that models with explicit spatial grounding objectives exhibit higher EICS, though our cross-architecture results suggest the failure is not easily corrected by objective alone.

\paragraph{Text-only probing.}
EICS is applied here to the \emph{text encoder} only. We probe hidden states at the \texttt{[EOS]} token, with mean pooling, and at the predicate's own tokens. We do not probe image encoder representations or joint vision--language representations. Our interventions are counterfactual edits to the caption; the image is never changed, so EICS measures how the text representation responds to a changed description, not how a model responds to a changed scene. Whether the image encoder encodes spatial arrangement more faithfully than the text encoder is a natural extension not addressed here.

\paragraph{Multi-sentence captions.}
CASP-VL captions render up to five scene-graph relationships, and a sentence-level representation may legitimately compress some per-sentence detail. Part of the spatial loss in our main results comes from this: with only the edited sentence, CLIP keeps 52\% of the BoW signal for spatial relations instead of 18\%, while color stays near BoW (98\%; Appendix~\ref{app:paraphrase}). The color--spatial gap therefore does not depend on multi-sentence input, but its size in the main results does.

\paragraph{Template captions.}
CASP-VL uses synthetically rendered template captions, several template sentences per image (``A red ball is on a blue table. \ldots''), which may differ in distribution from the natural image--caption pairs used to train CLIP. With human-written Visual Genome region descriptions or fluent LLM paraphrases in place of templates (Appendix~\ref{app:paraphrase}), CLIP keeps 97\% and 87\% of the lexical signal for color but only 31\% and 15\% for spatial relations. Template captions therefore lower the absolute scores without creating the gap.

\paragraph{Probe interpretability.}
EICS uses a logistic regression probe as its decoding mechanism. A near-zero EICS means the probe cannot track the intervention, but this does not rule out non-linear encoding. Our MLP probe comparison (Appendix~\ref{app:probe}) raises spatial EICS at L12 only from $+0.066$ to $+0.086$ (split 42), and at no layer to the linear BoW baseline, making a non-linearity explanation unlikely, but we cannot rule it out entirely for architectures we did not test.

% ── REFERENCES ───────────────────────────────────────────────────────────────
\bibliography{custom}

@inproceedings{
  yuksekgonul2023aro,
  title={When and why Vision-Language Models behave like  Bags-of-Words, and what to do about it?},
  author={Mert Yuksekgonul and Federico Bianchi and Pratyusha   Kalluri and Dan Jurafsky and James Zou},
  booktitle={International Conference on Learning Representations},
  year={2023},
  url={https://openreview.net/forum?id=KRLUvxh8uaX}
}

@article{hsieh2023sugarcrepe,
  title={{SugarCrepe}: Fixing hackable benchmarks for vision-language compositionality},
  author={Hsieh, Cheng-Yu and Zhang, Jieyu and Ma, Zixian and Kembhavi, Aniruddha and Krishna, Ranjay},
  journal={Advances in neural information processing systems},
  volume={36},
  pages={31096--31116},
  year={2023}
}

@article{zhao2024mmcomposition,
  title={{MMComposition}: Revisiting the compositionality of pre-trained vision-language models},
  author={Hua, Hang and Tang, Yunlong and Zeng, Ziyun and Cao, Liangliang and Yang, Zhengyuan and He, Hangfeng and Xu, Chenliang and Luo, Jiebo},
  journal={arXiv preprint arXiv:2410.09733},
  year={2024}
}

@inproceedings{thrush2022winoground,
  title={{Winoground}: Probing vision and language models for visio-linguistic compositionality},
  author={Thrush, Tristan and Jiang, Ryan and Bartolo, Max and Singh, Amanpreet and Williams, Adina and Kiela, Douwe and Ross, Candace},
  booktitle={2022 IEEE/CVF Conference on Computer Vision and Pattern Recognition (CVPR)},
  pages={5228--5238},
  year={2022},
  organization={IEEE}
}

@inproceedings{kamath2023whatsup,
  title={What’s “up” with vision-language models? investigating their struggle with spatial reasoning},
  author={Kamath, Amita and Hessel, Jack and Chang, Kai-Wei},
  booktitle={Proceedings of the 2023 Conference on Empirical Methods in Natural Language Processing},
  pages={9161--9175},
  year={2023}
}

@inproceedings{chen2024spatialvlm,
  title={{SpatialVLM}: Endowing vision-language models with spatial reasoning capabilities},
  author={Chen, Boyuan and Xu, Zhuo and Kirmani, Sean and Ichter, Brain and Sadigh, Dorsa and Guibas, Leonidas and Xia, Fei},
  booktitle={Proceedings of the IEEE/CVF conference on computer vision and pattern recognition},
  pages={14455--14465},
  year={2024}
}

@inproceedings{chen2025spatial_attention,
author = {Chen, Shiqi and Zhu, Tongyao and Zhou, Ruochen and Zhang, Jinghan and Gao, Siyang and Niebles, Juan Carlos and Geva, Mor and He, Junxian and Wu, Jiajun and Li, Manling},
title = {Why is spatial reasoning hard for {VLMs}? an attention mechanism perspective on focus areas},
year = {2025},
publisher = {JMLR.org},
booktitle = {Proceedings of the 42nd International Conference on Machine Learning},
articleno = {380},
numpages = {23},
location = {Vancouver, Canada},
series = {ICML'25}
}

@article{zhang2025cfvlm,
  title={{CF-VLM}: {CounterFactual} Vision-Language Fine-tuning},
  author={Cai, Kaitong and Fan, Yijia and Wang, Jian and Wang, Keze and others},
  journal={Advances in Neural Information Processing Systems},
  volume={38},
  pages={58056--58094},
  year={2026}
}

@InProceedings{parascandolo2025cogt,
        title={Causal Graphical Models for Vision-Language Compositional Understanding},
        author={Parascandolo, Fiorenzo and Moratelli, Nicholas and Sangineto, Enver and Baraldi, Lorenzo and Cucchiara, Rita},
        booktitle={Proceedings of The Thirteenth International Conference on Learning Representations, ICLR},
        year={2025}
    }

@article{yang2025exploring,
  title={Exploring information processing in large language models: Insights from information bottleneck theory},
  author={Yang, Zhou and Qi, Zhengyu and Ren, Zhaochun and Jia, Zhikai and Sun, Haizhou and Zhu, Xiaofei and Liao, Xiangwen},
  journal={arXiv preprint arXiv:2501.00999},
  year={2025}
}

@article{wani2025same,
  title={Same Answer, Different Representations: Hidden instability in {VLMs}},
  author={Wani, Farooq Ahmad and Suglia, Alessandro and Saxena, Rohit and Gema, Aryo Pradipta and Kwan, Wai-Chung and Barez, Fazl and Bucarelli, Maria Sofia and Silvestri, Fabrizio and Minervini, Pasquale},
  journal={arXiv preprint arXiv:2602.06652},
  year={2026}
}

@article{belinkov2022probing,
  title={Probing classifiers: Promises, shortcomings, and advances},
  author={Belinkov, Yonatan},
  journal={Computational Linguistics},
  volume={48},
  number={1},
  pages={207--219},
  year={2022}
}

@inproceedings{kang2026linear,
  title={Linear mechanisms for spatiotemporal reasoning in vision language models},
  author={Kang, Raphaela and Chen, Hongqiao and Gkioxari, Georgia and Perona, Pietro},
  booktitle={International Conference on Learning Representations},
  volume={2026},
  pages={41245--41258},
  year={2026}
}

@inproceedings{shani2026tokens,
  title={From tokens to thoughts: How {LLMs} and humans trade compression for meaning},
  author={Shani, Chen and Soffer, Liron and Jurafsky, Dan and LeCun, Yann and Shwartz-Ziv, Ravid},
  booktitle={International Conference on Learning Representations},
  volume={2026},
  pages={83019--83055},
  year={2026}
}

@article{tishby2000information,
  title={The information bottleneck method},
  author={Tishby, Naftali and Pereira, Fernando C and Bialek, William},
  journal={arXiv preprint physics/0004057},
  year={2000}
}

@inproceedings{radford2021clip,
  title={Learning transferable visual models from natural language supervision},
  author={Radford, Alec and Kim, Jong Wook and Hallacy, Chris and Ramesh, Aditya and Goh, Gabriel and Agarwal, Sandhini and Sastry, Girish and Askell, Amanda and Mishkin, Pamela and Clark, Jack and others},
  booktitle={International conference on machine learning},
  pages={8748--8763},
  year={2021},
  organization={PmLR}
}

@inproceedings{zhai2023siglip,
  title={Sigmoid loss for language image pre-training},
  author={Zhai, Xiaohua and Mustafa, Basil and Kolesnikov, Alexander and Beyer, Lucas},
  booktitle={2023 IEEE/CVF International Conference on Computer Vision (ICCV)},
  pages={11941--11952},
  year={2023},
  organization={IEEE}
}

@inproceedings{liu2024llava,
  title={Improved baselines with visual instruction tuning},
  author={Liu, Haotian and Li, Chunyuan and Li, Yuheng and Lee, Yong Jae},
  booktitle={2024 IEEE/CVF Conference on Computer Vision and Pattern Recognition (CVPR)},
  pages={26286--26296},
  year={2024},
  organization={IEEE}
}

@article{krishna2017visualgenome,
  title={Visual {Genome}: Connecting language and vision using crowdsourced dense image annotations},
  author={Krishna, Ranjay and Zhu, Yuke and Groth, Oliver and Johnson, Justin and Hata, Kenji and Kravitz, Joshua and Chen, Stephanie and Kalantidis, Yannis and Li, Li-Jia and Shamma, David A and others},
  journal={International journal of computer vision},
  volume={123},
  number={1},
  pages={32--73},
  year={2017},
  publisher={Springer}
}

@article{qwen2024,
  title={{Qwen2.5} Technical Report},
  author={{Qwen Team}},
  journal={arXiv preprint arXiv:2412.15115},
  year={2024}
}

@article{jeong2026cs,
  title={{CS-CLIP}: Compositional Scene Graph-guided {CLIP} for Robust Compositional Reasoning},
  author={Jeong, SeongJun and Jung, Minjoon and Choi, Woo Suk and Jang, Youwon and Zhang, Byoung-Tak},
  journal={arXiv preprint arXiv:2609.08242},
  year={2026}
}

@inproceedings{li2026cross,
  title={Cross-Modal Masked Compositional Concept Modeling for Enhancing Visio-Linguistic Compositionality},
  author={Li, Wei and Huang, Zhen and Tian, Xinmei},
  booktitle={Proceedings of the 64th Annual Meeting of the Association for Computational Linguistics (Volume 1: Long Papers)},
  pages={32284--32308},
  year={2026}
}

@inproceedings{herzig2023incorporating,
  title={Incorporating Structured Representations into Pretrained Vision \& Language Models Using Scene Graphs},
  author={Herzig, Roei and Mendelson, Alon and Karlinsky, Leonid and Arbelle, Assaf and Feris, Rogerio and Darrell, Trevor and Globerson, Amir},
  booktitle={Proceedings of the 2023 Conference on Empirical Methods in Natural Language Processing},
  pages={14077--14098},
  year={2023}
}

% ── APPENDIX ─────────────────────────────────────────────────────────────────
% A reader's FAQ first; then sections grouped by topic: data and protocol, robustness checks, implementation.
\appendix

\section{Frequently Asked Questions}
\label{app:faq}

Short answers to the objections we expect, each pointing to the analysis that addresses it. Unless noted, numbers are CLIP ViT-B/32 at L12 \texttt{[EOS]} under the main protocol.

\paragraph{Q1. What does an EICS of $+0.07$ mean, and what is the maximum?}
EICS is accuracy on the new label minus chance ($1/12$ for color, $1/16$ for spatial relations), so $0$ is chance and the ceiling is $0.917$ (color) or $0.938$ (spatial). The informative comparison is with BoW on the same split: CLIP keeps 94\% of the BoW signal for color but 19\% for spatial relations (\S\ref{sec:rq2}).

\paragraph{Q2. Why compare against a bag of words?}
BoW measures what is already in the input tokens. The edited word is in the caption, so a contextual encoder should keep at least what word counts keep. For spatial relations, CLIP keeps far less (Appendix~\ref{app:bow}).

\paragraph{Q3. Is this an artifact of template captions?}
No. With human-written Visual Genome region descriptions, CLIP keeps 97\% of the BoW signal for color and 31\% for spatial relations; with fluent Qwen2.5 paraphrases, 87\% and 15\% (Appendix~\ref{app:paraphrase}).

\paragraph{Q4. Is this just five sentences being compressed into one vector?}
Partly. With only the edited sentence, spatial EICS rises from $+0.066$ to $+0.487$, but CLIP still keeps only 52\% of the BoW signal for spatial relations, against 98\% for color (Appendix~\ref{app:paraphrase}). The gap does not depend on multi-sentence input; its size in the main results does.

\paragraph{Q5. Could the relation be encoded non-linearly?}
An MLP probe raises spatial EICS at L12 only from $+0.066$ to $+0.086$ (split 42), and at no layer does it reach the linear BoW baseline (Appendix~\ref{app:probe}).

\paragraph{Q6. Is the text encoder too small?}
Widening it does not help: spatial EICS is $+0.066$, $+0.053$, and $+0.045$ for ViT-B/32, ViT-L/14, and ViT-L/14@336 (Appendix~\ref{app:larger}).

\paragraph{Q7. Does CLIP see the relation word at all?}
Yes. Pooling over the predicate's own tokens gives $+0.930$, close to the ceiling. The relation is identifiable where it is written and largely lost in \texttt{[EOS]} and in mean pooling (\S\ref{sec:rq3}).

\paragraph{Q8. Does one lucky split drive the result?}
No. Across five image-grouped splits CLIP keeps 81--102\% of the BoW signal for color and 16--22\% for spatial relations (Table~\ref{tab:seeds}, Appendix~\ref{app:reeval}).

\paragraph{Q9. Some perturbed captions still contain the old label. Does that explain the gap?}
It works against our conclusion, not for it. Without such captions, color EICS ($+0.391$) is above BoW ($+0.326$), while spatial EICS stays at a fifth of it ($+0.074$ vs.\ $+0.366$; Appendix~\ref{app:errors}).

\paragraph{Q10. Does CLIP encode who is where?}
Weakly. With words fixed and arguments swapped, a probe tells the two captions apart with $0.574$ accuracy (BoW: $0.500$), and part of this may reflect plausibility rather than order (\S\ref{sec:rq3}).

\paragraph{Q11. CLIP scores 69\% on SugarCrepe relations. Isn't that a contradiction?}
No. Benchmarks ask whether the model prefers the right caption; EICS asks whether the representation tracks the change. A model can pass the first while keeping little of the relation (\S\ref{sec:rq4}).

\paragraph{Q12. Does the image fix it?}
No. After a spatial edit, the image prefers the original caption in only 34\% of cases (Appendix~\ref{app:grounding}).

\paragraph{Q13. Why do some tables use a different split?}
Only the first BoW robustness check (Appendix~\ref{app:bow}) uses an earlier triplet-level split, which keeps unscorable flips and lowers absolute values. Every other result uses the main protocol, and every CLIP-versus-BoW comparison points the same way under both protocols (Appendix~\ref{app:repro}). Counting CLIP's BPE tokens instead of words changes BoW by at most $0.012$ (Appendix~\ref{app:bow}).

\section{Cosine Similarity Under Concept Interventions}
\label{app:cosine}

Table~\ref{tab:cosine} reports the mean cosine similarity between the \texttt{[EOS]} hidden states
of original and perturbed captions in CLIP ViT-B/32, over all 2{,}400 color and 1{,}491 relation
triplets (\S\ref{sec:rq1}).

\begin{table}[h]
\centering
\caption{\textbf{Relation flips barely change CLIP's text representation, even at the last layer.}
Mean cosine similarity between original and perturbed \texttt{[EOS]} states (CLIP ViT-B/32).
Values near 1.0 mean the representation barely responds to the edit.}
\label{tab:cosine}
\small
\begin{tabular}{lccc}
\toprule
\rowcolor{CLRheader}
Flip & Embedding layer & L6 & L12 \\
\midrule
Color            & 1.000 & 0.997 & 0.943 \\
\rowcolor{CLRbad}Relation & 0.996 & 0.996 & \textbf{0.992} \\
\bottomrule
\end{tabular}
\end{table}

\section{Full Benchmark Results}
\label{app:benchmarks}

Table~\ref{tab:benchmarks} lists every ARO and SugarCrepe subtask for the three contrastive models,
computed with the same scoring script as Table~\ref{tab:sota} (accuracy over all examples).

\begin{table*}[t]
\centering
\caption{\textbf{SigLIP beats CLIP ViT-B/32 by 5--14 points on every attribute and object subtask, but not on the two relational ones ($+$1.3 on \texttt{replace\_rel}, $-$1.2 on \texttt{swap\_obj}).}
Accuracy (\%) on every ARO and SugarCrepe subtask (two-way choice, chance $=50\%$). Shaded rows
are the relational subtasks used in Table~\ref{tab:sota}.}
\label{tab:benchmarks}
\small
\begin{tabular}{llccc}
\toprule
\rowcolor{CLRheader}
Benchmark & Subtask & CLIP B/32 & CLIP L/14 & SigLIP B/16 \\
\midrule
\multirow{2}{*}{ARO} & VG-Relation    & 51.8 & 53.3 & 46.3 \\
                     & VG-Attribution & 61.4 & 61.6 & 54.0 \\
\midrule
\multirow{8}{*}{SugarCrepe} & \texttt{add\_att}     & 69.3 & 71.2 & 83.7 \\
                            & \texttt{replace\_att} & 80.5 & 79.1 & 87.2 \\
                            & \texttt{swap\_att}    & 63.9 & 61.9 & 71.6 \\
                            & \texttt{add\_obj}     & 77.0 & 78.5 & 89.1 \\
                            & \texttt{replace\_obj} & 90.6 & 94.0 & 95.4 \\
\rowcolor{CLRbad}           & \texttt{replace\_rel} & 69.3 & 65.3 & 70.6 \\
\rowcolor{CLRbad}           & \texttt{swap\_obj}    & 61.5 & 61.1 & 60.3 \\
                            & Overall               & 76.6 & 76.8 & 83.8 \\
\bottomrule
\end{tabular}
\end{table*}

\begin{table*}[t]
\centering
\caption{\textbf{The color--spatial gap is clear for the three contrastive models; LLaVA keeps little of either concept.}
EICS with 95\% bootstrap CIs at the layers of Table~\ref{tab:sota}, main protocol (split 42), with each model's BoW on the same split. SigLIP's 64-token limit removes a few triplets, so its split differs slightly (483 color, 239 spatial test triplets; 482 and 242 otherwise).}
\label{tab:models_ci}
\small
\begin{tabular}{lcccc}
\toprule
\rowcolor{CLRheader}
Model & Color EICS [95\% CI] & BoW & Spatial EICS [95\% CI] & BoW \\
\midrule
\rowcolor{CLRrow}CLIP ViT-B/32 & $+$0.292 [$+$0.251, $+$0.336] & $+$0.359 & $+$0.066 [$+$0.024, $+$0.107] & $+$0.359 \\
CLIP ViT-L/14                  & $+$0.303 [$+$0.259, $+$0.346] & $+$0.359 & $+$0.053 [$+$0.016, $+$0.090] & $+$0.359 \\
\rowcolor{CLRrow}SigLIP-B/16   & $+$0.339 [$+$0.296, $+$0.383] & $+$0.323 & $+$0.038 [$+$0.000, $+$0.080] & $+$0.331 \\
LLaVA-1.5-7B (L31)             & $+$0.143 [$+$0.107, $+$0.182] & $+$0.359 & $+$0.115 [$+$0.070, $+$0.165] & $+$0.359 \\
\bottomrule
\end{tabular}
\end{table*}

\section{Dataset Statistics}
\label{app:stats}

Table~\ref{tab:stats} summarizes CASP-VL, and Table~\ref{tab:flips} lists every edit in it.
Most captions render five scene-graph relationships; a third of them repeat a sentence because
Visual Genome annotates the same relationship more than once. Every relation edit changes exactly
one sentence, while 19\% of color edits change several, because the recolored object is mentioned
more than once. Two new relation labels, \emph{far from} and \emph{away from}, never occur as
original labels; the 300 flips that produce them (20\% of relation flips) cannot be scored by a
probe trained on original labels and are excluded from the main protocol
(Appendix~\ref{app:reeval}).

\begin{table}[h]
\centering
\caption{\textbf{CASP-VL at a glance.} Computed from the released triplet file with
\texttt{dataset\_stats.py}.}
\label{tab:stats}
\small
\setlength{\tabcolsep}{2.5pt}
\begin{tabular}{lcc}
\toprule
\rowcolor{CLRheader}
& Color & Relation \\
\midrule
Triplets                                  & 2{,}400 & 1{,}491 \\
Images                                    & 1{,}964 & 1{,}265 \\
Triplets per image (mean/max)& 1.22 / 5 & 1.18 / 5 \\
Original classes                          & 12 & 16 \\
Classes reached by edits                  & 8 & 15 \\
Captions with 5 sentences                 & 96.9\% & 96.0\% \\
Words per caption (mean, range)& 37.8 (7--52) & 37.3 (6--59) \\
Captions repeating a sentence& 34.4\% & 31.4\% \\
Edits changing $>$1 sentence& 19.1\% & 0\% \\
Old label left in $\tilde x$& 35.8\% & 36.9\% \\
Unedited copy of edited sentence& 12.3\% & 15.4\% \\
Originals shared by $>$1 triplet& 441 & 226 \\
Flips to a never-original label& 0 & 300 \\
\bottomrule
\end{tabular}
\end{table}

\begin{table*}[t]
\centering
\caption{\textbf{Every edit in CASP-VL, with counts.} Replacements are drawn from a fixed list of
alternatives per original label.}
\label{tab:flips}
\scriptsize
\setlength{\tabcolsep}{3pt}
\begin{tabular}{p{0.47\textwidth}p{0.47\textwidth}}
\toprule
\rowcolor{CLRheader}
Color flips (2{,}400) & Relation flips (1{,}491) \\
\midrule
pink$\to$red 106, grey$\to$white 106, purple$\to$blue 106, gray$\to$black 105, brown$\to$black 104,
white$\to$gray 104, black$\to$white 102, orange$\to$yellow 101, orange$\to$red 99, black$\to$brown 98,
white$\to$black 96, brown$\to$white 96, gray$\to$white 95, pink$\to$white 94, grey$\to$black 94,
purple$\to$green 94, yellow$\to$red 82, red$\to$green 78, green$\to$red 73, blue$\to$yellow 69,
green$\to$yellow 68, blue$\to$green 66, blue$\to$red 65, red$\to$yellow 65, yellow$\to$green 63,
green$\to$blue 59, red$\to$blue 57, yellow$\to$blue 55
&
near$\to$far from 100, in front of$\to$behind 100, inside$\to$outside 100, outside$\to$inside 100,
beside$\to$away from 100, next to$\to$far from 100, behind$\to$in front of 100, left of$\to$right of 63,
on top of$\to$beneath 55, below$\to$over 55, beneath$\to$on top of 54, above$\to$under 52,
over$\to$under 51, over$\to$below 49, above$\to$below 48, beneath$\to$on 46, below$\to$above 45,
on top of$\to$under 45, under$\to$above 39, on$\to$under 38, under$\to$over 36, on$\to$below 32,
on$\to$beneath 30, right of$\to$left of 28, under$\to$on 25 \\
\bottomrule
\end{tabular}
\end{table*}

\section{Main Evaluation Protocol}
\label{app:reeval}

The CLIP ViT-B/32 results in Section~\ref{sec:experiments} (Figure~\ref{fig:layerwise},
Table~\ref{tab:predtoken}) use the following protocol. The probe split is grouped by Visual Genome
image, so no image contributes triplets to both training and test, and no training original
caption appears verbatim as a test perturbed caption. Hidden states come from the Hugging Face CLIP
ViT-B/32 text encoder with its causal attention mask, and EICS is scored against the new label
$\tilde y_i$. Test flips whose new label never occurs as an original label in training are
excluded, because the probe cannot predict an unseen class: these are one-directional antonym flips
such as \emph{near} $\rightarrow$ \emph{far from} and \emph{next to} $\rightarrow$ \emph{away
from} (53 of 295 spatial test triplets). This leaves 1{,}918 training and 482 test color triplets
(12 classes) and 1{,}196 training and 242 test spatial triplets (16 trained relation classes).
Captions render several relations from the image's scene graph, for example
\emph{``A walking giraffe is near a standing ostrich. A railing is around a tree. A dirt path is on
a grass. \ldots''}, and the edit changes one of them (here \emph{near} $\rightarrow$ \emph{far from}).

\begin{table*}[t]
\centering
\caption{\textbf{Split 42 of Table~\ref{tab:predtoken}, with confidence intervals.}
EICS with 95\% bootstrap CIs over the 482 color and 242 spatial held-out triplets (* = CI above zero); number of classes in
parentheses. \emph{Orig.\ acc}: probe accuracy on original captions. Every single-split analysis in the appendix uses this split.}
\label{tab:reeval}
\small
\begin{tabular}{lcc}
\toprule
\rowcolor{CLRheader}
CLIP B/32, L12, split 42 & Color (12) & Spatial (16) \\
\midrule
\texttt{[EOS]}    & \cellcolor{CLRgood}$+$0.292 [$+$0.251, $+$0.336]* & \cellcolor{CLRbad}$+$0.066 [$+$0.024, $+$0.107]* \\
Mean pooling      & \cellcolor{CLRgood}$+$0.328 [$+$0.282, $+$0.371]* & \cellcolor{CLRbad}$+$0.095 [$+$0.049, $+$0.136]* \\
Predicate tokens  & $+$0.917 [$+$0.917, $+$0.917]* & $+$0.933 [$+$0.925, $+$0.938]* \\
\rowcolor{CLRbow}BoW & $+$0.359 [$+$0.315, $+$0.400]* & $+$0.359 [$+$0.297, $+$0.421]* \\
\midrule
Orig.\ acc, CLIP/BoW & 0.65 / 0.66 & 0.21 / 0.64 \\
\bottomrule
\end{tabular}
\end{table*}

\paragraph{Split stability.} Table~\ref{tab:seeds} repeats the main run on five image-grouped splits. Split 42, used for every single-split analysis, is the least favorable split for color (CLIP keeps 81\% of the BoW signal, against 92--102\% on the other four); for spatial relations all five splits agree (16--22\%).

\begin{table*}[t]
\centering
\caption{\textbf{The color--spatial gap holds on every split.} CLIP ViT-B/32, L12 \texttt{[EOS]}, main protocol, one row per split seed. All EICS values have their 95\% CI above zero. \emph{Kept}: CLIP EICS as a share of BoW EICS. Spatial test sets differ in size because unscorable flips are dropped per split.}
\label{tab:seeds}
\small
\setlength{\tabcolsep}{3.5pt}
\begin{tabular}{lcccccccc}
\toprule
\rowcolor{CLRheader}
& \multicolumn{4}{c}{Color} & \multicolumn{4}{c}{Spatial} \\
\cmidrule(lr){2-5}\cmidrule(lr){6-9}
\rowcolor{CLRheader}
Seed & $n$ & CLIP & BoW & Kept & $n$ & CLIP & BoW & Kept \\
\midrule
42 & 482 & $+$0.292 & $+$0.359 & 81\% & 242 & $+$0.066 & $+$0.359 & 18\% \\
1  & 498 & $+$0.292 & $+$0.316 & 92\% & 227 & $+$0.070 & $+$0.400 & 17\% \\
2  & 498 & $+$0.316 & $+$0.324 & 98\% & 218 & $+$0.070 & $+$0.364 & 19\% \\
3  & 487 & $+$0.327 & $+$0.334 & 98\% & 234 & $+$0.087 & $+$0.391 & 22\% \\
4  & 484 & $+$0.334 & $+$0.328 & 102\% & 247 & $+$0.055 & $+$0.346 & 16\% \\
\midrule
\rowcolor{CLRrow}Mean & & $+$0.312 & $+$0.332 & 94\% & & $+$0.070 & $+$0.372 & 19\% \\
\rowcolor{CLRrow}s.d. & & 0.019 & 0.016 & & & 0.012 & 0.022 & \\
\bottomrule
\end{tabular}
\end{table*}

\paragraph{Synonyms and polysemy.} Merging near-synonymous predicates
(\{above, over\}, \{below, beneath, under\}, \{on, on top of\}) gives spatial EICS of $+0.227$
[$+0.165$, $+0.289$] for CLIP against $+0.375$ [$+0.313$, $+0.442$] for BoW, and excluding the
polysemous \emph{on} and \emph{over} gives $+0.120$ against $+0.305$. The gap is therefore not an
artifact of near-synonym labels or polysemous predicates. No flip in the data exchanges two
predicates from the same synonym group.

\paragraph{Object priors.} On perturbed captions, both probes still predict the \emph{old}
label more often than chance (spatial: by $+0.08$ for CLIP and $+0.12$ for BoW; color: $+0.11$ and
$+0.09$), because the subject and object nouns, which carry information about the likely relation
or attribute, are unchanged by the edit. This pulls every EICS value below its ceiling, BoW's
included.

\section{Error Analysis}
\label{app:errors}

Tables~\ref{tab:conf_color} and~\ref{tab:conf_spatial} break the L12 \texttt{[EOS]} probe's predictions on perturbed captions down by the new label (CLIP ViT-B/32, main protocol, seed 42; the same matrices for every model and seed are in the released result files).

\begin{table*}[t]
\centering
\caption{\textbf{Color: most perturbed captions move toward the new color, and errors spread over neighboring colors.} Confusion of the L12 \texttt{[EOS]} probe on perturbed captions (CLIP ViT-B/32, main protocol, seed 42). Rows: the new label $\tilde y$; columns: predicted class; cells: counts, shaded by share of the row (green = correct, red = error). \emph{new/old/other \%}: share predicted as the new label, the original label, or anything else.}
\label{tab:conf_color}
\footnotesize
\setlength{\tabcolsep}{2.2pt}
\renewcommand{\arraystretch}{1.15}
\begin{tabular}{l@{\hspace{5pt}}cccccccccccc@{\hspace{6pt}}rrrr}
\toprule
new label $\tilde y$ & \rotatebox{90}{black} & \rotatebox{90}{blue} & \rotatebox{90}{brown} & \rotatebox{90}{gray} & \rotatebox{90}{green} & \rotatebox{90}{grey} & \rotatebox{90}{orange} & \rotatebox{90}{pink} & \rotatebox{90}{purple} & \rotatebox{90}{red} & \rotatebox{90}{white} & \rotatebox{90}{yellow} & $n$ & \rotatebox{90}{new \%} & \rotatebox{90}{old \%} & \rotatebox{90}{other \%} \\
\midrule
black & \cellcolor{CLRheatok!26}\textbf{20} & \cellcolor{CLRheaterr!8}6 & \cellcolor{CLRheaterr!8}5 & \cellcolor{CLRheaterr!8}2 & \cellcolor{CLRheaterr!14}11 & \cellcolor{CLRheaterr!8}1 & \cellcolor{CLRheaterr!8}4 & \cellcolor{CLRheaterr!8}1 & \cellcolor{CLRheaterr!8}2 & \cellcolor{CLRheaterr!8}4 & \cellcolor{CLRheaterr!9}7 & \cellcolor{CLRheaterr!8}3 & 66 & 30 & 17 & 53 \\
blue & \cellcolor{CLRheaterr!8}2 & \cellcolor{CLRheatok!45}\textbf{28} & \cellcolor{CLRheaterr!8}4 & \textcolor{black!25}{0} & \textcolor{black!25}{0} & \textcolor{black!25}{0} & \cellcolor{CLRheaterr!8}1 & \cellcolor{CLRheaterr!8}2 & \cellcolor{CLRheaterr!8}5 & \cellcolor{CLRheaterr!10}6 & \cellcolor{CLRheaterr!8}3 & \cellcolor{CLRheaterr!8}2 & 53 & 53 & 11 & 36 \\
brown & \cellcolor{CLRheaterr!27}5 & \cellcolor{CLRheaterr!8}1 & \cellcolor{CLRheatok!16}\textbf{3} & \textcolor{black!25}{0} & \cellcolor{CLRheaterr!8}1 & \textcolor{black!25}{0} & \cellcolor{CLRheaterr!8}1 & \textcolor{black!25}{0} & \textcolor{black!25}{0} & \cellcolor{CLRheaterr!11}2 & \cellcolor{CLRheaterr!11}2 & \cellcolor{CLRheaterr!8}1 & 16 & 19 & 31 & 50 \\
gray & \cellcolor{CLRheaterr!11}4 & \cellcolor{CLRheaterr!8}3 & \cellcolor{CLRheaterr!8}2 & \cellcolor{CLRheatok!16}\textbf{6} & \cellcolor{CLRheaterr!8}2 & \cellcolor{CLRheaterr!14}5 & \textcolor{black!25}{0} & \cellcolor{CLRheaterr!8}1 & \cellcolor{CLRheaterr!8}2 & \textcolor{black!25}{0} & \cellcolor{CLRheaterr!14}5 & \cellcolor{CLRheaterr!8}1 & 31 & 19 & 16 & 65 \\
green & \textcolor{black!25}{0} & \cellcolor{CLRheaterr!9}6 & \cellcolor{CLRheaterr!8}1 & \cellcolor{CLRheaterr!8}1 & \cellcolor{CLRheatok!39}\textbf{25} & \textcolor{black!25}{0} & \cellcolor{CLRheaterr!8}1 & \cellcolor{CLRheaterr!8}1 & \cellcolor{CLRheaterr!8}3 & \cellcolor{CLRheaterr!8}4 & \cellcolor{CLRheaterr!8}2 & \cellcolor{CLRheaterr!17}11 & 55 & 45 & 22 & 33 \\
red & \cellcolor{CLRheaterr!8}2 & \cellcolor{CLRheaterr!8}4 & \cellcolor{CLRheaterr!8}4 & \cellcolor{CLRheaterr!8}3 & \cellcolor{CLRheaterr!8}6 & \textcolor{black!25}{0} & \cellcolor{CLRheaterr!10}10 & \cellcolor{CLRheaterr!8}8 & \cellcolor{CLRheaterr!8}2 & \cellcolor{CLRheatok!39}\textbf{41} & \cellcolor{CLRheaterr!8}5 & \cellcolor{CLRheaterr!8}4 & 89 & 46 & 24 & 30 \\
white & \cellcolor{CLRheaterr!8}11 & \cellcolor{CLRheaterr!8}5 & \cellcolor{CLRheaterr!15}20 & \cellcolor{CLRheaterr!8}9 & \cellcolor{CLRheaterr!12}16 & \cellcolor{CLRheaterr!8}4 & \cellcolor{CLRheaterr!8}3 & \cellcolor{CLRheaterr!8}9 & \cellcolor{CLRheaterr!8}2 & \cellcolor{CLRheaterr!8}10 & \cellcolor{CLRheatok!18}\textbf{24} & \cellcolor{CLRheaterr!8}2 & 115 & 21 & 24 & 55 \\
yellow & \cellcolor{CLRheaterr!8}5 & \cellcolor{CLRheaterr!8}4 & \textcolor{black!25}{0} & \cellcolor{CLRheaterr!8}1 & \cellcolor{CLRheaterr!8}5 & \textcolor{black!25}{0} & \cellcolor{CLRheaterr!8}1 & \cellcolor{CLRheaterr!8}1 & \textcolor{black!25}{0} & \cellcolor{CLRheaterr!8}2 & \cellcolor{CLRheaterr!8}4 & \cellcolor{CLRheatok!51}\textbf{34} & 57 & 60 & 12 & 28 \\
\midrule
\multicolumn{13}{l}{all test triplets} & 482 & 38 & 20 & 43 \\
\bottomrule
\end{tabular}
\end{table*}

\begin{table*}[t]
\centering
\caption{\textbf{Spatial: there is no diagonal; the probe rarely reaches the new relation and mostly ignores the edit.} Same layout as Table~\ref{tab:conf_color}. Only relations that occur as a scorable new label in the test split are shown as rows; columns are the 16 classes of the probe.}
\label{tab:conf_spatial}
\footnotesize
\setlength{\tabcolsep}{2.2pt}
\renewcommand{\arraystretch}{1.15}
\begin{tabular}{l@{\hspace{5pt}}cccccccccccccccc@{\hspace{6pt}}rrrr}
\toprule
new label $\tilde y$ & \rotatebox{90}{above} & \rotatebox{90}{behind} & \rotatebox{90}{below} & \rotatebox{90}{beneath} & \rotatebox{90}{beside} & \rotatebox{90}{in front of} & \rotatebox{90}{inside} & \rotatebox{90}{left of} & \rotatebox{90}{near} & \rotatebox{90}{next to} & \rotatebox{90}{on} & \rotatebox{90}{on top of} & \rotatebox{90}{outside} & \rotatebox{90}{over} & \rotatebox{90}{right of} & \rotatebox{90}{under} & $n$ & \rotatebox{90}{new \%} & \rotatebox{90}{old \%} & \rotatebox{90}{other \%} \\
\midrule
above & \cellcolor{CLRheatok!18}\textbf{4} & \cellcolor{CLRheaterr!8}1 & \cellcolor{CLRheaterr!8}1 & \cellcolor{CLRheaterr!18}4 & \textcolor{black!25}{0} & \textcolor{black!25}{0} & \textcolor{black!25}{0} & \textcolor{black!25}{0} & \textcolor{black!25}{0} & \textcolor{black!25}{0} & \cellcolor{CLRheaterr!13}3 & \cellcolor{CLRheaterr!9}2 & \cellcolor{CLRheaterr!8}1 & \cellcolor{CLRheaterr!9}2 & \textcolor{black!25}{0} & \cellcolor{CLRheaterr!8}1 & 19 & 21 & 11 & 68 \\
behind & \cellcolor{CLRheaterr!8}1 & \cellcolor{CLRheatok!24}\textbf{6} & \textcolor{black!25}{0} & \textcolor{black!25}{0} & \cellcolor{CLRheaterr!8}2 & \cellcolor{CLRheaterr!12}3 & \cellcolor{CLRheaterr!8}1 & \textcolor{black!25}{0} & \textcolor{black!25}{0} & \cellcolor{CLRheaterr!8}1 & \cellcolor{CLRheaterr!8}1 & \textcolor{black!25}{0} & \cellcolor{CLRheaterr!20}5 & \cellcolor{CLRheaterr!8}1 & \textcolor{black!25}{0} & \textcolor{black!25}{0} & 21 & 29 & 14 & 57 \\
below & \cellcolor{CLRheaterr!8}3 & \cellcolor{CLRheaterr!8}2 & \textcolor{black!25}{0} & \cellcolor{CLRheaterr!11}4 & \cellcolor{CLRheaterr!8}1 & \textcolor{black!25}{0} & \cellcolor{CLRheaterr!11}4 & \textcolor{black!25}{0} & \cellcolor{CLRheaterr!8}2 & \textcolor{black!25}{0} & \cellcolor{CLRheaterr!8}1 & \cellcolor{CLRheaterr!8}1 & \cellcolor{CLRheaterr!14}5 & \cellcolor{CLRheaterr!16}6 & \textcolor{black!25}{0} & \cellcolor{CLRheaterr!8}2 & 31 & 0 & 6 & 94 \\
beneath & \textcolor{black!25}{0} & \cellcolor{CLRheaterr!11}2 & \textcolor{black!25}{0} & \textcolor{black!25}{0} & \cellcolor{CLRheaterr!11}2 & \textcolor{black!25}{0} & \cellcolor{CLRheaterr!8}1 & \textcolor{black!25}{0} & \cellcolor{CLRheaterr!8}1 & \cellcolor{CLRheaterr!8}1 & \textcolor{black!25}{0} & \cellcolor{CLRheaterr!21}4 & \cellcolor{CLRheaterr!11}2 & \cellcolor{CLRheaterr!8}1 & \textcolor{black!25}{0} & \cellcolor{CLRheaterr!11}2 & 16 & 0 & 12 & 88 \\
in front of & \textcolor{black!25}{0} & \cellcolor{CLRheaterr!13}3 & \textcolor{black!25}{0} & \textcolor{black!25}{0} & \textcolor{black!25}{0} & \cellcolor{CLRheatok!18}\textbf{4} & \cellcolor{CLRheaterr!13}3 & \textcolor{black!25}{0} & \cellcolor{CLRheaterr!9}2 & \cellcolor{CLRheaterr!8}1 & \cellcolor{CLRheaterr!9}2 & \textcolor{black!25}{0} & \cellcolor{CLRheaterr!9}2 & \cellcolor{CLRheaterr!8}1 & \textcolor{black!25}{0} & \cellcolor{CLRheaterr!8}1 & 19 & 21 & 16 & 63 \\
inside & \cellcolor{CLRheaterr!8}1 & \cellcolor{CLRheaterr!8}1 & \textcolor{black!25}{0} & \textcolor{black!25}{0} & \cellcolor{CLRheaterr!8}1 & \cellcolor{CLRheaterr!8}1 & \cellcolor{CLRheatok!15}\textbf{3} & \textcolor{black!25}{0} & \textcolor{black!25}{0} & \textcolor{black!25}{0} & \cellcolor{CLRheaterr!8}1 & \cellcolor{CLRheaterr!10}2 & \cellcolor{CLRheaterr!30}6 & \textcolor{black!25}{0} & \textcolor{black!25}{0} & \cellcolor{CLRheaterr!8}1 & 17 & 18 & 35 & 47 \\
on & \cellcolor{CLRheaterr!8}1 & \cellcolor{CLRheaterr!8}1 & \cellcolor{CLRheaterr!10}2 & \textcolor{black!25}{0} & \cellcolor{CLRheaterr!10}2 & \cellcolor{CLRheaterr!15}3 & \textcolor{black!25}{0} & \textcolor{black!25}{0} & \cellcolor{CLRheaterr!8}1 & \cellcolor{CLRheaterr!8}1 & \cellcolor{CLRheatok!8}\textbf{1} & \cellcolor{CLRheaterr!10}2 & \textcolor{black!25}{0} & \textcolor{black!25}{0} & \textcolor{black!25}{0} & \cellcolor{CLRheaterr!15}3 & 17 & 6 & 12 & 82 \\
on top of & \cellcolor{CLRheaterr!8}1 & \textcolor{black!25}{0} & \cellcolor{CLRheaterr!8}1 & \textcolor{black!25}{0} & \cellcolor{CLRheaterr!12}2 & \textcolor{black!25}{0} & \textcolor{black!25}{0} & \textcolor{black!25}{0} & \cellcolor{CLRheaterr!12}2 & \cellcolor{CLRheaterr!8}1 & \cellcolor{CLRheaterr!12}2 & \cellcolor{CLRheatok!18}\textbf{3} & \cellcolor{CLRheaterr!8}1 & \textcolor{black!25}{0} & \textcolor{black!25}{0} & \cellcolor{CLRheaterr!8}1 & 14 & 21 & 0 & 79 \\
outside & \textcolor{black!25}{0} & \textcolor{black!25}{0} & \textcolor{black!25}{0} & \textcolor{black!25}{0} & \cellcolor{CLRheaterr!8}1 & \cellcolor{CLRheaterr!11}2 & \cellcolor{CLRheaterr!42}8 & \textcolor{black!25}{0} & \textcolor{black!25}{0} & \textcolor{black!25}{0} & \textcolor{black!25}{0} & \cellcolor{CLRheaterr!11}2 & \cellcolor{CLRheatok!11}\textbf{2} & \textcolor{black!25}{0} & \textcolor{black!25}{0} & \cellcolor{CLRheaterr!8}1 & 16 & 12 & 50 & 38 \\
over & \cellcolor{CLRheaterr!9}2 & \cellcolor{CLRheaterr!13}3 & \cellcolor{CLRheaterr!8}1 & \textcolor{black!25}{0} & \cellcolor{CLRheaterr!9}2 & \textcolor{black!25}{0} & \cellcolor{CLRheaterr!8}1 & \textcolor{black!25}{0} & \cellcolor{CLRheaterr!9}2 & \textcolor{black!25}{0} & \cellcolor{CLRheaterr!9}2 & \cellcolor{CLRheaterr!8}1 & \cellcolor{CLRheaterr!8}1 & \cellcolor{CLRheatok!9}\textbf{2} & \textcolor{black!25}{0} & \cellcolor{CLRheaterr!9}2 & 19 & 11 & 5 & 84 \\
right of & \textcolor{black!25}{0} & \cellcolor{CLRheaterr!8}1 & \textcolor{black!25}{0} & \textcolor{black!25}{0} & \cellcolor{CLRheaterr!17}2 & \textcolor{black!25}{0} & \cellcolor{CLRheaterr!8}1 & \textcolor{black!25}{0} & \textcolor{black!25}{0} & \cellcolor{CLRheaterr!8}1 & \cellcolor{CLRheaterr!34}4 & \textcolor{black!25}{0} & \textcolor{black!25}{0} & \cellcolor{CLRheaterr!8}1 & \textcolor{black!25}{0} & \textcolor{black!25}{0} & 10 & 0 & 0 & 100 \\
under & \cellcolor{CLRheaterr!8}2 & \cellcolor{CLRheaterr!8}2 & \cellcolor{CLRheaterr!8}2 & \cellcolor{CLRheaterr!8}2 & \textcolor{black!25}{0} & \cellcolor{CLRheaterr!8}3 & \cellcolor{CLRheaterr!10}5 & \textcolor{black!25}{0} & \cellcolor{CLRheaterr!8}3 & \cellcolor{CLRheaterr!8}3 & \cellcolor{CLRheaterr!10}5 & \cellcolor{CLRheaterr!8}4 & \cellcolor{CLRheaterr!8}3 & \cellcolor{CLRheaterr!8}3 & \textcolor{black!25}{0} & \cellcolor{CLRheatok!12}\textbf{6} & 43 & 14 & 14 & 72 \\
\midrule
\multicolumn{17}{l}{all test triplets} & 242 & 13 & 14 & 73 \\
\bottomrule
\end{tabular}
\end{table*}

\paragraph{Color moves, spatial does not.} For color, 38\% of perturbed captions are assigned the new label, 20\% keep the original label, and 43\% go elsewhere; the diagonal is the largest cell in most rows. For spatial relations there is no diagonal: 13\% reach the new relation, 14\% keep the old one, and 73\% go elsewhere. The edit usually does not change the prediction at all: the probe gives the same answer to the original and the perturbed caption for 86\% of spatial triplets (208/242), against 51\% for color (244/482). Conditioning on originals the probe gets right makes the contrast sharper. Of 313 such color triplets, 140 move to the new color; of the 50 such spatial triplets, 4 move to the new relation and 35 keep the old one. When CLIP has identified the relation, replacing it is mostly invisible at \texttt{[EOS]}.

\paragraph{Spatial errors are mostly not near-misses.} If CLIP encoded spatial relations coarsely, errors would fall on a synonym of the new relation (\emph{over}/\emph{above}, \emph{under}/\emph{beneath}). They spread across most columns instead: no row puts half of its mass on a single column, except the \emph{outside} row, half of which keeps the old label \emph{inside}. Synonyms do absorb part of the error: merging the synonym groups raises spatial EICS from $+0.066$ to $+0.227$, but BoW rises with it to $+0.375$, so CLIP still keeps only 60\% of the lexical signal (Appendix~\ref{app:reeval}).

\paragraph{Examples.} Real test triplets from this run (one edited sentence of each caption shown verbatim; the probe's prediction for original $\to$ perturbed):

\begin{tcolorbox}[colback=CLRfindbg,colframe=CLRfind,boxrule=0.4pt,arc=1pt,left=4pt,right=4pt,top=3pt,bottom=3pt,fontupper=\small]
\textbf{Spatial, edit ignored} \hfill probe: \textit{outside} $\to$ \textit{outside}\\
\hspace*{6pt}\emph{``A light is outside a shop.''}\\
\hspace*{6pt}$\to$ \emph{``A light is inside a shop.''}\par\smallskip
\textbf{Spatial, edit tracked} \hfill probe: \textit{behind} $\to$ \textit{in front of}\\
\hspace*{6pt}\emph{``A tree is behind a tan cow.''}\\
\hspace*{6pt}$\to$ \emph{``A tree is in front of a tan cow.''}\par\smallskip
\textbf{Color, edit tracked} \hfill probe: \textit{yellow} $\to$ \textit{green}\\
\hspace*{6pt}\emph{``A young kid is WEARING a yellow shirt.''}\\
\hspace*{6pt}$\to$ \emph{``A young kid is WEARING a green shirt.''}\par\smallskip
\textbf{Color, edit ignored} \hfill probe: \textit{gray} $\to$ \textit{gray}\\
\hspace*{6pt}\emph{``A gray cover is under a bar.''}\\
\hspace*{6pt}$\to$ \emph{``A white cover is under a bar.''}
\end{tcolorbox}

\paragraph{Leftover old labels explain most color errors, not spatial ones.} A color edit recolors one object instance. Visual Genome often annotates the same relationship several times with different object instances, so the perturbed caption can keep an unedited copy of the edited sentence (\emph{``\ldots a red building. \ldots a pink building.''}), and other objects may share the old color. Overall, the old label still occurs somewhere in 35.8\% of perturbed color captions and 36.9\% of perturbed relation captions (Table~\ref{tab:stats}). Table~\ref{tab:leftover} splits the test set by this property. Color EICS is $+0.433$ when the old color is gone from the perturbed caption and $+0.051$ when it is still present, where half of the predictions keep the old label. Spatial EICS is low either way ($+0.062$ and $+0.072$). To test this directly, we drop every triplet whose perturbed caption still contains the old label (858 color, 550 relation triplets) and rerun the full protocol, probe and BoW alike. On the remaining 1,542 color and 941 relation triplets, CLIP's color EICS ($+0.391$) is above BoW ($+0.326$), while spatial EICS stays at a fifth of it ($+0.074$ vs.\ $+0.366$; Table~\ref{tab:leftover}). Predicate-span pooling still recovers the relation ($+0.882$). Leftover labels therefore depress color EICS and leave spatial EICS unchanged: the headline gap is conservative.

\begin{table*}[t]
\centering
\caption{\textbf{Without leftover old labels, CLIP's color EICS exceeds BoW, while spatial EICS stays at a fifth of it.} Top: test triplets of the main run (CLIP ViT-B/32, L12 \texttt{[EOS]}) split by whether the old label still occurs in the perturbed caption (point estimates; \emph{old \%}: share predicted as the old label). Bottom: probe and BoW retrained and scored on the leftover-free data only, with 95\% bootstrap CIs.}
\label{tab:leftover}
\small
\setlength{\tabcolsep}{3.5pt}
\begin{tabular}{llcccc}
\toprule
\rowcolor{CLRheader}
Concept & Old label in $\tilde x$? & $n$ & CLIP L12 & BoW & old \% \\
\midrule
\multicolumn{6}{l}{\emph{Main run, test set split}} \\
\rowcolor{CLRgood}Color   & no  & 304 & $+$0.433 & & 2 \\
\rowcolor{CLRgood}        & yes & 178 & $+$0.051 & & 51 \\
\rowcolor{CLRbad}Spatial  & no  & 160 & $+$0.062 & & 9 \\
\rowcolor{CLRbad}         & yes &  82 & $+$0.072 & & 26 \\
\midrule
\multicolumn{6}{l}{\emph{Retrained on leftover-free data}} \\
\rowcolor{CLRgood}Color   & no & 310 & $+$0.391 {\footnotesize[+0.339, +0.446]}* & $+$0.326 {\footnotesize[+0.271, +0.381]}* & \\
\rowcolor{CLRbad}Spatial  & no & 161 & $+$0.074 {\footnotesize[+0.018, +0.124]}* & $+$0.366 {\footnotesize[+0.291, +0.447]}* & \\
\bottomrule
\end{tabular}
\end{table*}

\section{BoW Robustness Check}
\label{app:bow}

The BoW baseline applies the identical EICS probe to
\emph{normalized word-count vectors} rather than CLIP hidden states. Each
caption is represented as a vector of word counts, $\ell_2$-normalized. Words
are split on word boundaries (not CLIP's BPE), so a multi-word predicate such as
\emph{on top of} contributes three words, one of which (\emph{on}) it shares
with another class. A logistic
regression probe trained on BoW vectors of original captions is then evaluated
on BoW vectors of perturbed captions. If BoW EICS exceeds CLIP EICS, CLIP's
12-layer contextual encoding has actively removed discriminative information
that was present in the raw token surface form.

This comparison is meaningful because token identity \emph{does} change under
both color flips (``red'' $\to$ ``blue'') and spatial flips
(``on'' $\to$ ``beneath''). A model that encoded even shallow lexical
information faithfully would achieve EICS comparable to BoW. The fact that
CLIP spatial EICS falls far below BoW spatial EICS (under the triplet-level split, $+$0.040
vs.\ $+$0.173; under the main protocol, $+$0.066 vs.\ $+$0.359) is therefore not a failure to learn something hard; it is a
failure to preserve something trivially present in the input.

Counting CLIP's BPE tokens instead of words gives the same baseline under the main protocol: $+0.354$ instead of $+0.359$ for color and $+0.347$ instead of $+0.359$ for spatial relations (split 42), so the choice of tokenization does not favor multi-word predicates.

\section{Linear vs.\ MLP Probe Comparison}
\label{app:probe}
Linear probing is a standard tool for measuring what information is linearly decodable from a representation~\citep{belinkov2022probing}. Spatial relations could, however, be encoded non-linearly. We therefore rerun the main protocol with an MLP probe in place of logistic regression (scikit-learn \texttt{MLPClassifier}: two hidden layers of 256 and 128 ReLU units, $\alpha=10^{-4}$, early stopping on a 20\% validation split of the training originals with patience 10, \texttt{random\_state}=42). The BoW baseline is probed with the same MLP, so both sides of the comparison change together.

\begin{table}[h]
\centering
\caption{\textbf{An MLP probe raises early-layer spatial EICS, but at no layer does it reach the linear BoW baseline.} EICS by layer for a linear and an MLP probe (CLIP ViT-B/32, main protocol, seed 42, \texttt{[EOS]}). Bottom row: BoW with each probe.}
\label{tab:exp5}
\small
\setlength{\tabcolsep}{4.5pt}
\begin{tabular}{lcccc}
\toprule
\rowcolor{CLRheader}
& \multicolumn{2}{c}{Color} & \multicolumn{2}{c}{Spatial} \\
\cmidrule(lr){2-3}\cmidrule(lr){4-5}
\rowcolor{CLRheader}
Layer & Linear & MLP & Linear & MLP \\
\midrule
L00 & $-$0.009 & $-$0.021 & $-$0.001 & $+$0.020 \\
L02 & $+$0.039 & $+$0.099 & $+$0.045 & $+$0.214 \\
L04 & $+$0.056 & $+$0.120 & $+$0.111 & $+$0.177 \\
L06 & $+$0.025 & $+$0.074 & $+$0.107 & $+$0.165 \\
L08 & $+$0.035 & $+$0.049 & $+$0.070 & $+$0.111 \\
L10 & $+$0.027 & $+$0.072 & $+$0.057 & $+$0.053 \\
L11 & $+$0.263 & $+$0.244 & $+$0.049 & $+$0.061 \\
L12 & $+$0.292 & $+$0.319 & $+$0.066 & $+$0.086 \\
\midrule
\rowcolor{CLRbow}BoW & $+$0.359 & $+$0.259 & $+$0.359 & $+$0.218 \\
\bottomrule
\end{tabular}
\end{table}

Table~\ref{tab:exp5} shows the result. At the last layer the MLP changes little: color rises from $+0.292$ to $+0.319$ and spatial from $+0.066$ to $+0.086$. In the early layers, where the representation is still close to the token embeddings, the MLP extracts more spatial signal (up to $+0.214$ at L02), roughly what the same MLP extracts from BoW ($+0.218$); by L10 this advantage is gone. Under neither probe does spatial EICS at any layer reach the linear BoW baseline ($+0.359$). The MLP also does worse than logistic regression on the sparse BoW vectors, so a more flexible probe is not uniformly a stronger one. The late-layer failure is therefore not an artifact of the linear probe: together with predicate-token probing (\S\ref{sec:rq3}), the result points to the relation being lost in \texttt{[EOS]}, not hidden in a non-linear code.

\section{Visual Grounding}
\label{app:grounding}

To test whether the image supplies the spatial information missing from the
text representation, we compute
$\Delta_i = \mathrm{sim}(\mathrm{img}_i, x_i) - \mathrm{sim}(\mathrm{img}_i, \tilde{x}_i)$
for up to 500 triplets (250 attribute and 250 relation flips, keeping those whose image is
available), using CLIP ViT-B/32's projected image and text embeddings, where $x_i$ and $\tilde{x}_i$ are the original and perturbed
captions. Positive $\Delta$ means the image prefers the original caption.
For spatial relations, grounding \emph{amplifies} the failure: images prefer
the original only 34.0\% of the time ($\Delta=-0.0014$;
Table~\ref{tab:exp2}). One reading is that the image encoder captures which objects
co-occur more than how they are arranged.

\begin{table}[h]
\centering
\caption{\textbf{Adding the image helps for color and size but hurts for spatial relations: after a spatial flip, the image prefers the wrong caption 66\% of the time.}
Image--text alignment delta (CIs are $t$-intervals). ``Img $\succ$ orig'' is the fraction of triplets where the
image prefers the original caption.}
\label{tab:exp2}
\small
\setlength{\tabcolsep}{3.5pt}
\begin{tabular}{lcc}
\toprule
\rowcolor{CLRheader}
Intervention & $\Delta$ mean [95\% CI] & Img $\succ$ orig \\
\midrule
Color flip   & $+$0.0040 $[+0.003, +0.005]$ & 61.6\% \\
Size flip    & $+$0.0057 $[+0.001, +0.010]$ & 55.9\% \\
\rowcolor{CLRbad}Spatial flip & $-$0.0014 $[-0.002, -0.001]$ & 34.0\% \\
\bottomrule
\end{tabular}
\end{table}

\section{Relation Taxonomy Analysis}
\label{app:taxonomy}

We ask whether spatial, part-whole, and action relations are
representationally distinguishable using 4,200 cross-category triplets
(Table~\ref{tab:exp6}). A majority-class baseline correction rules out
probe bias.

\begin{table*}[t]
\centering
\caption{\textbf{Swapping another relation into a spatial one is detectable, but swapping a spatial relation out is not: spatial predicates dominate the geometry.}
Cross-category EICS at L12 ($n=4{,}200$).
\emph{Flip rate}: fraction of perturbed captions on which the probe
predicts the new (perturbed) label; EICS subtracts chance ($1/|C|=0.333$).
\emph{Baseline}: probe's spatial prediction rate on unperturbed
source-class captions. \emph{Corrected}: for edits \emph{into} the spatial class, flip rate minus
this baseline; for edits \emph{out of} it, flip rate minus chance (1/3).}
\label{tab:exp6}
\small
\begin{tabular}{lcccc}
\toprule
\rowcolor{CLRheader}
Transition & Flip rate & Baseline & Corrected & 95\% CI \\
\midrule
spatial $\to$ part-whole & 0.018 & 0.932 & $-$0.315 & [$-$0.341, $-$0.289] \\
spatial $\to$ action     & 0.087 & 0.932 & $-$0.246 & [$-$0.276, $-$0.216] \\
part-whole $\to$ spatial & 0.808 & 0.675 & $+$0.133 & [$+$0.094, $+$0.172] \\
action $\to$ spatial     & 0.438 & 0.278 & $+$0.160 & [$+$0.112, $+$0.208] \\
\bottomrule
\end{tabular}
\end{table*}

The probe predicts \emph{spatial} for 67.5\% of unperturbed part-whole
captions, confirming part-whole and action relations occupy the same
representational region. Edits out of the spatial class almost never move the prediction: after
spatial $\to$ part-whole the probe predicts \emph{part-whole} for only 1.8\% of perturbed
captions, far below chance. Edits into it do: after part-whole $\to$ spatial it predicts
\emph{spatial} 13.3 points more often than it already does on unperturbed part-whole captions. The geometry is dominated by spatial predicates to the
point where other relation types are invisible to linear probes.

\section{Caption-Style Robustness}
\label{app:paraphrase}

CASP-VL captions are template sentences (``A walking giraffe is near a standing ostrich. \ldots''), which CLIP did not see in training. Low spatial EICS could therefore reflect a distribution mismatch rather than a failure to encode the relation. We test this by treating caption style as a manipulated variable and keeping everything else fixed: same images, same flips, same main protocol (image-grouped split, seed 42, CLIP ViT-B/32, \texttt{[EOS]} at L12, BoW baseline).

\paragraph{VG region descriptions (human-written).} For each triplet we take the human-written Visual Genome region descriptions~\citep{krishna2017visualgenome} of the \emph{same image} and keep those that contain the edited word as a whole word exactly once (so \emph{on} never matches inside \emph{person} or \emph{on top of}). We then drop any description that also names another label of the same kind (e.g., \emph{``red and yellow stripe''} for a red$\to$blue flip), so the edited word is the only cue. One description is drawn at random (seed 42) as the original; the perturbed caption replaces only the edited word (\emph{``bus in front of a wall''} $\to$ \emph{``bus behind a wall''}). Of the 3,891 triplets, 440 have no such description and 146 have only descriptions with a competing label, leaving 3,305 (2,204 color, 1,101 relation). The description need not mention the edited object itself; the probe target is the label word, as in the main data.

\paragraph{Single sentence.} Each CASP-VL caption renders up to five scene-graph relationships, and a pooled representation of five sentences may legitimately compress some per-sentence detail. We therefore also keep only the first sentence that the edit changed, in both the original and the perturbed caption (\emph{``A walking giraffe is near a standing ostrich.''} $\to$ \emph{``\ldots is far from a standing ostrich.''}). All 3,891 triplets qualify. In 207 color triplets this sentence still names the old color elsewhere (e.g., \emph{``a red yellow shirt''}); a second run drops them, together with every triplet whose perturbed caption contains the old label (Appendix~\ref{app:errors}).

\paragraph{LLM paraphrases.} We ask Qwen2.5-7B-Instruct~\citep{qwen2024} to rewrite each template caption into fluent English (greedy decoding). The edited occurrences are marked in the input as \texttt{[[word]]} and the model is asked to copy them unchanged; the perturbed caption is the rewrite with only the marked words replaced. A rewrite is kept only if it keeps every mark and the same count of every other label word, so the two captions still differ in exactly one concept. Of the 3,891 triplets, 133 are not a clean word swap in the template, 1,104 rewrites lost or altered a marked word, 453 added or dropped another label word, and 19 were unchanged or kept stray brackets, leaving 2,182 (1,500 color, 682 relation). The \emph{left of} and \emph{right of} classes fall below the 20-example minimum in this subset, so the relation probe has 14 classes.

\begin{table*}[t]
\centering
\caption{\textbf{Under every caption format, CLIP keeps nearly all of the color signal but loses much of the spatial signal; one sentence per caption narrows the spatial loss without closing it.} EICS at L12 (\texttt{[EOS]}) under the main protocol; $n$ = test triplets. BoW reaches its ceiling ($1-1/|C|$) on VG regions and single spatial sentences because each names exactly one label of its kind. * = 95\% bootstrap CI above zero.}
\label{tab:paraphrase}
\small
\setlength{\tabcolsep}{4pt}
\begin{tabular}{llcccc}
\toprule
\rowcolor{CLRheader}
Style & Concept & $n$ & CLIP L12 & BoW & CLIP/BoW \\
\midrule
\rowcolor{CLRrow}Template   & Color   & 482 & $+$0.292* & $+$0.359* & 81\% \\
\rowcolor{CLRrow}           & Spatial & 242 & $+$0.066* & $+$0.359* & 18\% \\
VG regions & Color   & 448 & $+$0.890* & $+$0.917* & 97\% \\
           & Spatial & 184 & $+$0.291* & $+$0.938* & 31\% \\
\rowcolor{CLRrow}LLM paraphrase & Color   & 300 & $+$0.370* & $+$0.427* & 87\% \\
\rowcolor{CLRrow}               & Spatial & 110 & $+$0.065* & $+$0.438* & 15\% \\
Single sentence & Color   & 482 & $+$0.765* & $+$0.784* & 98\% \\
                & Spatial & 242 & $+$0.487* & $+$0.938* & 52\% \\
\rowcolor{CLRrow}Single, no leftovers & Color   & 310 & $+$0.768* & $+$0.800* & 96\% \\
\rowcolor{CLRrow}                     & Spatial & 161 & $+$0.428* & $+$0.875* & 49\% \\
\bottomrule
\end{tabular}
\end{table*}

\paragraph{Results.} Table~\ref{tab:paraphrase} reports the results; CLIP's 95\% intervals are $[+0.874,+0.903]$ (color) and $[+0.220,+0.361]$ (spatial) on VG regions, and $[+0.310,+0.423]$ and $[+0.010,+0.129]$ on LLM paraphrases. Short human-written descriptions raise all scores, because a single phrase gives the probe one label and little competing content. The gap between concepts does not close: CLIP keeps 97\% of the lexical signal for color but only 31\% for spatial relations, and its spatial probe classifies only 58.7\% of the \emph{original} descriptions correctly, against 98.0\% for color. Predicate-span pooling again recovers the relation ($+0.872$, against a ceiling of $0.938$), so the relation word is identifiable at its own tokens but is not carried into \texttt{[EOS]}. Fluent LLM paraphrases of the same scenes reproduce the template result almost exactly: CLIP keeps 87\% of the lexical signal for color and 15\% for spatial relations (template: 81\% and 18\%), and predicate-span pooling again recovers the relation ($+0.892$). Natural and human-written captions therefore reproduce the template finding; templates make the absolute scores lower but do not create the gap.

With a single sentence per caption, the gap narrows but does not close. CLIP keeps 98\% of the lexical signal for color ($+0.765$ [$+0.732$, $+0.796$] vs.\ $+0.784$) but only 52\% for spatial relations ($+0.487$ [$+0.429$, $+0.545$] vs.\ $+0.938$), and 49\% when leftover old labels are also removed. Pooling five sentences therefore accounts for part of the spatial loss in the main results (where CLIP keeps 18\%), but even one sentence loses about half of the relation at \texttt{[EOS]}. Mean pooling keeps more of a single sentence ($+0.702$), and predicate-token pooling again reaches the ceiling ($+0.938$): the relation is fully identifiable at its own tokens and partly lost when the sentence is summarized.

\section{Model Size Scaling}
\label{app:larger}

An obvious alternative explanation for near-zero spatial EICS in CLIP
ViT-B/32 is insufficient model capacity: a 512-dimensional text encoder may
simply lack the representational volume to encode spatial relations. We test
this by repeating the L12 EICS analysis on two larger CLIP variants:
ViT-L/14 (768-dimensional text encoder, trained on the same 400M
image--text pairs as ViT-B/32~\citep{radford2021clip}) and ViT-L/14@336
(identical architecture, fine-tuned at 336px image resolution). The text
encoder depth (12 layers) is identical across all three variants; only the
hidden dimension changes. If capacity were limiting, we would expect EICS to
increase with dimension.

We run the main protocol (image-grouped split, seed 42, L12 \texttt{[EOS]}) for each model. If BoW approximates what a wider encoder could keep, spatial EICS should move toward BoW as width grows.

\begin{table*}[t]
\centering
\caption{\textbf{A wider text encoder does not raise spatial EICS, so the failure is not a capacity limit.} EICS at L12 \texttt{[EOS]} under the main protocol (482 color, 242 spatial test triplets; the BoW baseline does not depend on the model). * = 95\% CI above zero.}
\label{tab:larger}
\small
\setlength{\tabcolsep}{3pt}
\begin{tabular}{lccc}
\toprule
\rowcolor{CLRheader}
Model & Width & Color EICS [CI] & Spatial EICS [CI] \\
\midrule
\rowcolor{CLRrow}ViT-B/32     & 512 & $+$0.292 {\footnotesize[+0.251, +0.336]*} & $+$0.066 {\footnotesize[+0.024, +0.107]*} \\
ViT-L/14     & 768 & $+$0.303 {\footnotesize[+0.259, +0.346]*} & $+$0.053 {\footnotesize[+0.016, +0.090]*} \\
\rowcolor{CLRrow}ViT-L/14@336 & 768 & $+$0.298 {\footnotesize[+0.257, +0.344]*} & $+$0.045 {\footnotesize[+0.008, +0.086]*} \\
\midrule
\rowcolor{CLRbow}BoW          & --- & $+$0.359 {\footnotesize[+0.315, +0.400]*} & $+$0.359 {\footnotesize[+0.297, +0.421]*} \\
\bottomrule
\end{tabular}
\end{table*}

Color EICS rises slightly with width ($+0.292$, $+0.303$, $+0.298$), while spatial EICS does not rise at all ($+0.066$, $+0.053$, $+0.045$), with overlapping confidence intervals (Table~\ref{tab:larger}). Both larger models also reach the predicate-span ceiling ($+0.933$), so the relation is again identifiable at its own tokens and lost at \texttt{[EOS]}. Widening the text encoder from 512 to 768 dimensions does not help spatial encoding: within the range tested, the failure is not a capacity bottleneck.

\section{Implementation and Reproducibility}
\label{app:repro}
\sloppy

\paragraph{Hardware and software.} All experiments ran on a single NVIDIA RTX A5500 GPU (24\,GB;
driver 596.71) under Windows with an Anaconda environment, using Python 3.13.5, PyTorch 2.10.0
(CUDA 12.8), Hugging Face Transformers 5.3.0, \texttt{open\_clip} 3.3.0, NumPy 2.1.3, and
scikit-learn 1.6.1. Encoding all
original and perturbed captions with a text encoder takes seconds on this GPU; probes run on the CPU.

\paragraph{Models.} Text encoders are loaded from the Hugging Face Hub:
\url{openai/clip-vit-base-patch32} (main model), \url{openai/clip-vit-large-patch14},
\url{google/siglip-base-patch16-224}, and \url{llava-hf/llava-1.5-7b-hf} (loaded in float16).
SigLIP text is padded to its training length (\texttt{padding="max\_length"}, 64 tokens); other
encoders use their default tokenizers. For CLIP we use the 13 hidden states returned by the text
encoder (embedding output and the 12 transformer blocks, before the final layer norm), computed with
the model's causal attention mask.

\paragraph{Representations.} \texttt{[EOS]} pooling takes the hidden state at the end-of-text
token (the position of token id 49407). Mean pooling averages all non-padding positions after the
start token, including \texttt{[EOS]}. Predicate pooling locates the predicate's BPE tokens in the
caption by exact subsequence match, using the original predicate $y_i$ for $x_i$ and the new
predicate $\tilde y_i$ for $\tilde x_i$, and averages their hidden states; the match succeeded for
every caption. All representations are $\ell_2$-normalized before probing. The BoW baseline
lowercases each caption, splits it into words with the regular expression \verb|\b\w+\b|, builds
its vocabulary from the training originals only, and $\ell_2$-normalizes the count vectors.

\paragraph{Probe, split, and statistics.} The probe is scikit-learn's \texttt{LogisticRegression}
($C=1.0$, \texttt{lbfgs}, \texttt{max\_iter}$=2000$, \texttt{random\_state}$=42$), trained on
$h_\ell(x_i)$ with labels $y_i$. The split shuffles the unique Visual Genome image ids with seed 42
and assigns the first 80\% of images, with all their triplets, to training. Test flips whose new
label is not among the training labels are dropped (Appendix~\ref{app:reeval}). EICS is the
fraction of test triplets on which $f_\ell(h_\ell(\tilde x_i)) = \tilde y_i$, minus $1/|\mathcal{C}|$
with $|\mathcal{C}|$ the number of training classes. Confidence intervals are percentile intervals
from 1{,}000 bootstrap resamples of the test triplets.

\paragraph{Triplet-level analyses.} The analyses marked ``triplet-level split'' use an 80/20
split over triplets stratified by the original label (scikit-learn \texttt{train\_test\_split},
seed 42), with the label set taken as the union of original and new labels, so chance is
$1/|\mathcal{C}|$ over that union and unscorable flips count as errors; probes use
\texttt{max\_iter}$=1000$ and otherwise the settings above. The triplet-level BoW baseline counts CLIP BPE
tokens (including start and end tokens) over the full 49{,}408-token vocabulary, $\ell_1$-normalized,
with the \texttt{saga} solver. The taxonomy analysis
(Appendix~\ref{app:taxonomy}) trains a three-way probe (spatial / part-whole / action) on L12
\texttt{[EOS]} states of 4{,}200 cross-category triplets.

\paragraph{LLaVA.} Each caption is fed to LLaVA-1.5-7B as \texttt{USER: <image>\textbackslash
n\{caption\}\textbackslash nASSISTANT:} with a blank $336\times336$ gray image, and we probe the
hidden state of the last prompt token at language-model layers 0, 4, 8, \ldots, 28, and 31 (of 32);
Table~\ref{tab:sota} reports the final layer, L31. Color EICS peaks at L28 ($+0.100$) and spatial
EICS at L4 ($+0.056$), so choosing each concept's best layer would not change any comparison.
We sample 100 color and 80 relation triplets per original class.

\paragraph{Benchmarks.} ARO VG-Relation accuracy is computed over all 23{,}937 examples (micro
average). SugarCrepe uses the seven official subtask files with COCO 2017 validation images; 1{,}549
of the 1{,}560 required images were available, and items whose image was missing were skipped. A
contrastive model is correct when the image scores the positive caption above the hard negative.

\paragraph{Data and code.} CASP-VL is released as one JSON file of 3{,}891 triplets (2{,}400 color
flips and 1{,}491 relation flips). Each triplet stores the Visual Genome image id, the original and
perturbed captions, and an edit log with the edit type and the old and new value. One script
reproduces all main-protocol CLIP results (run once per concept), and a second script regenerates
Figures~\ref{fig:eics_concept} and~\ref{fig:layerwise} from its output; both are included in the
release at \url{https://github.com/Shubhf/CASP-VL}.

\end{document}